\documentclass{article}

\usepackage[preprint, nonatbib]{neurips_2021}

\usepackage[utf8]{inputenc} 
\usepackage[T1]{fontenc}    
\usepackage{hyperref}       
\usepackage{url}            
\usepackage{booktabs}       
\usepackage{multirow}
\usepackage{amsfonts}       
\usepackage{amsmath}
\usepackage{nicefrac}       
\usepackage{microtype}      
\usepackage{xcolor}         
\usepackage[pdftex]{graphicx}
\usepackage{subfigure}
\usepackage[numbers,square,comma,sort,compress]{natbib}
\usepackage[ruled,linesnumbered]{algorithm2e}
\usepackage{pifont}
\usepackage{pythonhighlight}

\title{Scalable and Adaptive Graph Neural Networks with Self-Label-Enhanced Training}

\author{%
  Chuxiong Sun
  \\
  Emerging Technology Research Division \\
  China Telecom Research Institute \\
  Beijing, China\\
  \texttt{chuxiongsun@gmail.com} \\
   \And
  Hongming Gu\\
  Emerging Technology Research Division \\
  China Telecom Research Institute \\
  Beijing, China\\
  \texttt{guhm@chinatelecom.cn} \\
  \AND
  Jie Hu \\
  Emerging Technology Research Division \\
  China Telecom Research Institute \\
  Beijing, China\\
  \texttt{hujie1@chinatelecom.cn} \\
}

\begin{document}

\maketitle

\begin{abstract}
Besides of the existing neighbor sampling techniques applied on common Graph Neural Networks (GNNs), scalable methods allowing normal minibatch training can more easily scale to large scaled graphs. They decouple graph convolutions and other learnable transformations into preprocessing and a scalable classifier. A complex and graph structure-aware classifier is important to achieve competitive performances. By replacing redundant concatenation operation in Scalable Inception Graph Neural Networks (SIGN) with a more graph structure-aware attention mechanism, we propose Scalable and Adaptive Graph Neural Networks (SAGN). SAGN can adaptively gather neighborhood information among different hops. To further improve scalable GNNs by introducing the existing techniques applied on common GNNs for semi-supervised learning tasks, we propose Self-Label-Enhanced (SLE) training approach combining the self-training approach and label propagation in depth. We add the base model with a scalable label model. Then we iteratively train models and enhance the training set in several stages. To generate input of the label model, we apply label propagation based on one-hot encoded label vectors without inner random masking. We find out that empirically the label leakage has been effectively alleviated with enough propagation depth. The hard pseudo labels in the enhanced training set participate in label propagation with true labels, which propagates model knowledge and label information into the whole graph. Experiments on both inductive and transductive datasets demonstrate that, compared with other sampling-based and sampling-free methods, SAGN achieves better or comparable results and SLE can further improve performance.
\end{abstract}

\section{Introduction}

Common Graph Neural Networks (GNNs) have state-of-the-art performances on graph related tasks \cite{kipf2016semi, zhang2018link,qi2018learning,monti2017geometric,choma2018graph,duvenaud2015convolutional,gilmer2017neural,parisot2018disease,zitnik2018modeling,rossi2019ncrna,monti2019fake,gainza2020deciphering}. To scale them to large scaled graphs, neighbor sampling methods \cite{hamilton2017inductive,ying2018graph,huang2018adaptive,chen2018fastgcn,chen2017stochastic,zou2019layer,chiang2019cluster,zeng2019graphsaint} have been proposed. In another way, the scalable GNNs \cite{wu2019simplifying, rossi2020sign, huang2020combining} can more easily scale to large scaled graphs with better simplicity and runtime. These methods decouple graph convolutions and learnable transformations into preprocessing and a scalable classifier. They allow standard and efficient mini-batch training since the input is Euclidean. But non-learnable graph convolutions in preprocessing potentially limit expressiveness. Thus a complex and graph structure-aware classifier is required to achieve comparable performance. 

The self-training approach \cite{li2018deeper,sun2020multistage,yang2020self} and label propagation \cite{xiaojin2002learning,wang2007label,fujiwara2014efficient,wang2020unifying,shi2020masked} have been exploited on common GNNs for semi-supervised learning tasks. But there is a lack of research on incorporating these techniques into scalable GNNs, which may bring better performance in addition to simplicity and efficiency. The self-training approach can be easily applied to scalable GNNs. To incorporate label information while avoiding a trivial solution, the inner random masking within the training set \cite{shi2020masked, wang2021bag} is required for each epoch. However, for scalable GNNs, the associated label propagation for each epoch will cost much runtime. Moreover, the combination of self-training and label propagation has not been adequately discussed.



By designing a more expressive classifier, we propose \underline{\textbf{S}}calable and \underline{\textbf{A}}daptive \underline{\textbf{G}}raph neural \underline{\textbf{N}}etworks (\textbf{SAGN}). SAGN inherits inception-like module from SIGN and incorporates learnable attention weights between multiple hops \cite{sun2020adaptive}. The attention mechanism improves performance and interpretability. 
By effectively combining self-training and label propagation for scalable GNNs, we propose the \underline{\textbf{S}}elf-\underline{\textbf{L}}abel-\underline{\textbf{E}}nhanced (\textbf{SLE}) training approach. Firstly, to incorporate label information, we propose a scalable label model compatible with any scalable base model whose input is the propagated label embedding. We find that empirically the label leakage is effectively reduced with enough propagation number. Secondly, following the self-training approach, we split the training process into stages with enhancing the training set. Thirdly, the hard pseudo labels in self-training participate in label propagation at each stage.

\paragraph{Contributions}
In conclusion, we propose a new scalable GNN (SAGN) and associated training approach (SLE) for large scaled graphs on semi-supervised learning tasks. Our main contributions are: 1. We incorporate a learnable and structure-aware attention mechanism into scalable GNNs to obtain better expressiveness and interpretability with considerable complexity. 2. We first propose that empirically the label propagation can effectively alleviate the label leakage without inner random masking, which makes it feasible to incorporate label information into scalable GNNs. 3. We first propose a general training approach effectively combining self-training and label propagation to boost scalable GNNs on semi-supervised learning tasks.

\paragraph{Experiments}
We conduct experiments on both inductive and transductive datasets including the largest public node classification dataset. SAGN scales well to web-scale data. And SAGN achieves better results compared with other sampling-based and sampling-free baselines on most datasets. SLE can significantly boost performance of SAGN with relative improvements up to 8.4\%. On ogbn-products and ogbn-papers100M we obtain state-of-the-art results.

\section{Background and related works}
\paragraph{Graph neural networks}
The original spectral GNNs \cite{bruna2013spectral,kipf2016semi,defferrard2016convolutional} require complicate matrix eigendecomposition. Chebyshev Spectral CNN (ChebNet) uses a simplified approximation method \cite{defferrard2016convolutional,levie2019transferability,levie2018cayleynets}. Graph Convolution Network (GCN) \cite{kipf2016semi} further limits the approximation order and adds self-loops. The spatial GNNs, including GCN, Graph Attention Networks (GATs) \cite{velivckovic2017graph} and Graph Isomorphism Network (GIN) \cite{xu2018powerful}, perform direct neighborhood aggregations in graphs. In application scenarios, it is difficult for common GNNs to scale to huge graphs. Because the receptive field of sampled nodes is variable and grows exponentially with layers.

\paragraph{Sampling-based Methods}
The neighbor sampling methods for common GNNs have been proven to be effective to scale to huge graphs, which can be classified into node-wise, layer-wise and graph-wise methods. Node-wise methods \cite{hamilton2017inductive,ying2018graph,chen2017stochastic} sample a small number of neighbors for each node. Layer-wise methods \cite{chen2018fastgcn,huang2018adaptive} avoid the redundant sampling of node-wise methods. Graph-wise methods \cite{chiang2019cluster,zeng2019graphsaint} partition graph and limit convolutions in smaller subgraphs. However, they require complicate implementation and may still perform redundant convolution computation.

\begin{figure}[b]
    \centering
    \includegraphics[width=0.70\linewidth]{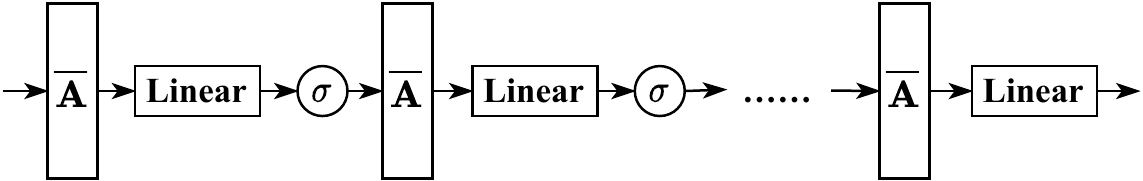}
    \caption{Architecture of common GNNs.}
    \label{fig: common GNNs}
\end{figure}



\paragraph{Scalable GNNs}

As shown in Figure \ref{fig: common GNNs}, common GNNs alternately stack graph convolutions, linear layers and non-linear activation functions to generate non-linear hierarchies. Scalable GNNs \cite{wu2019simplifying, rossi2020sign} decouple the model into preprocessing and a scalable classifier. 

Simplified GCN (SGC) \cite{wu2019simplifying} first points out that empirically the removal of intermediate non-linear activations does not significantly affect the model performance. SGC achieves comparable performances in small datasets with much faster runtime. However, its single linear layer limits expressiveness and the multi-hop information is lost.

SIGN \cite{rossi2020sign} uses more complex multi-layer perceptrons (MLP) and an inception-like module \cite{szegedy2015going}. Firstly, each aggregation is passed through its associated MLP encoder. Secondly, the encoded representations are concatenated and forwarded into a post MLP classifier to obtain final results. The multi-hop MLP encoders before concatenation can somehow approximate non-linear hierarchies of common GNNs. SIGN achieves significantly better performances than SGC. However, its concatenation operation is naive and brings extra memory cost, which is equal to summing with uniform weights. In another word, SIGN does not explicitly consider the importances of hops for different nodes, which are related to different local distributions.



\paragraph{Self-training approach on graphs}
For semi-supervised learning tasks, we can enhance the training set with more "pseudo-labeled" nodes to improve GNNs via the self-training approaches \cite{li2018deeper, yang2020self, sun2020multistage}. Because these extra nodes can contain valuable local information which does not appear in the raw training set. 
Li et al. \cite{li2018deeper} first propose the co-training and self-training approaches on GNNs. But it usually needs to train multiple models to obtain improvements. Self-Enhanced GNN \cite{yang2020self} incorporates topology update before enhancing the training set. It denoises the node relations by adding intra-class edges and removing inter-class edges based on predicted labels. However, low-quality predictions can remove valuable edges and add redundant edges. Sun et al. \cite{sun2020multistage} propose Multi-Stage Training Framework (M3S) combined with DeepCluster approach \cite{caron2018deep}. It performs DeepCluster on the trained node representations and aligns cluster labels with true labels to obtain pseudo labels. But it requires both high-quality predictions and clustering results.

\paragraph{Label propagation}
Besides of enhancing the training set, we can also perform label propagation to improve performance. A simple method is to concatenate node features and one-hot encoded label embeddings \cite{wang2021bag}. Shi et al. \cite{shi2020masked} add node feature matrix with transformed label embedding matrix. Wang et al. \cite{wang2020unifying} use label propagation prior in computing edge weights. Correct and Smooth (C\&S) \cite{huang2020combining} propagates residuals (or corrections) of training set nodes into unlabeled nodes and performs label smoothing \cite{klicpera2018predict}. Since C\&S is non-parameterized postprocessing, it can be viewed as a scalable method with graph convolutions as postprocessing. But for simple base models it requires complex preprocessing including spectral and diffusion \cite{klicpera2019diffusion} calculation, which may cost much preprocessing runtime.

\section{Proposed methods}


In this section, we firstly present a more expressive scalable model SAGN. Then we present the SLE training approach using SAGN (for example) as the base model, which combines self-training and label propagation.

\paragraph{Notations}
Let $\mathcal G$ represent a graph, $\mathcal N$ represent its node set, $\mathcal E$ represent its edge set, $X$ represent its node feature matrix, $C$ represent the number of classes, $\mathbf A$ represent its adjacency matrix, $N$ represent its node number, $E$ represent its edge number, $\mathcal L_s$ represent the enhanced training set at the $s$-stage, $\mathcal U_s$ represent the its complementary set, $Y$ represent the raw one-hot encoded label embedding, $\hat{Y}_{s}$ represent soft output labels of the $s$-th model and $\tilde{Y}_{s}$ represent hard output labels. Note that $Y$ is zero vector for nodes out of the raw training set.

\begin{figure}[tb]
    \centering
    \includegraphics[width=0.70\linewidth]{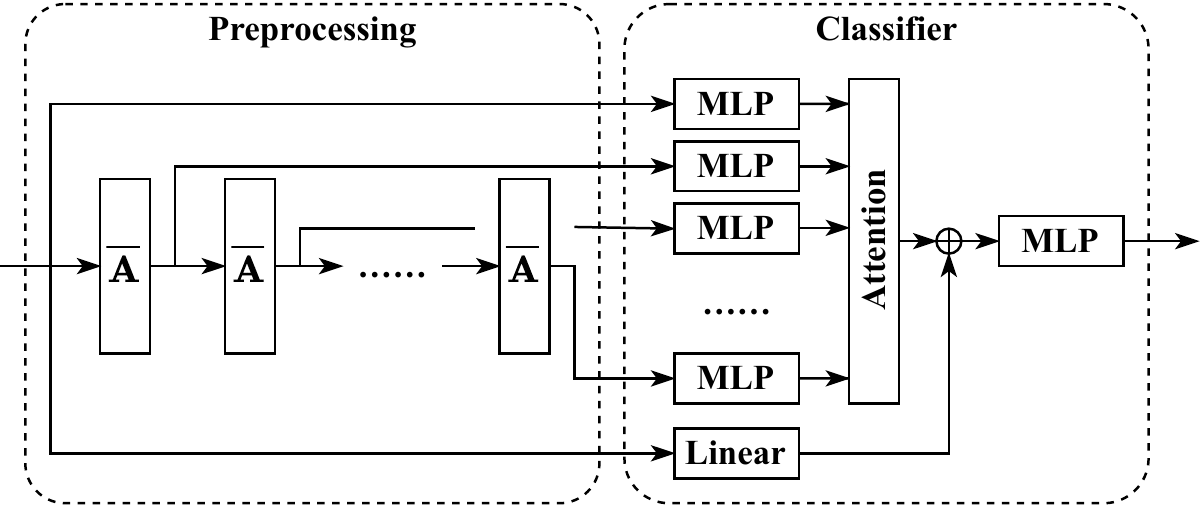}
    \caption{Architecture of SAGN. The multi-hop encoders and post encoder are depicted in form of MLP. The bottom "linear" refers to the residual linear layer. The symbol \textcircled{+} represents the operation of summing integrated representation and residual term. }
    \label{fig:SAGN}
\end{figure}

\subsection{SAGN}
As shown in Figure \ref{fig:SAGN}, SAGN follows the overall architecture of previous scalable models and incorporates an attention mechanism.

\paragraph{Preprocessing}

Suppose that we want to gather neighborhood node features within $K_f$-hop convolutions. We set $\overline{X}^{(0)}=X$. The $k$-hop smoothed node feature matrix $\overline {X}^{(k)}$ ($k>0$) is computed as below:
\begin{equation}
\label{eq: update smoothed node features}
\overline {X}^{(k)}=\overline{A}\overline {X}^{(k-1)},
\end{equation}
where $\overline{A}$ is the non-parametric transition matrix \cite{klicpera2019diffusion} which can be row-stochastic random walk matrix, symmetrically normalized adjacency matrix, etc.

\paragraph{Multi-hop encoders}
As shown in Figure \ref{fig:SAGN}, SAGN uses multi-hop node features as input. To obtain the $k$-hop representation $H^{(k)}$, we apply an MLP encoder $\zeta^{(k)}$ as below:
\begin{equation}
H^{(k)}=\zeta^{(k)}(\overline{X}^{(k)}).
\end{equation}

\paragraph{Attention mechanism}
Then we sum different encoded representations with diagonal attention matrices \cite{velivckovic2017graph, sun2020adaptive}. We hypothesise that multi-hop information varies among different nodes. We obtain the integrated representation $H_{att}$ as below:
\begin{equation}
H_{att}=\sum_{k=0}^{K_f}\Theta^{(k)}H^{(k)},
\end{equation}
where $\Theta^{(k)}$ is the $k$-th diagonal attention matrix. The $i$-th entry $\theta^{(k)}_i$ of $\Theta^{(k)}$ is calculated as below:
\begin{equation}
\theta^{(k)}_i=\mathop{\rm softmax}_k(\mathop{\rm LeakyReLU}([H^{(0)}_i||H^{(k)}_i]\cdot a)),
\end{equation}
where $a$ is the attention vector, $||$ represents concatenation operation and $\cdot$ represents dot product.

\paragraph{Post encoder}
With adding a residual term, we further feed the integrated representation into the post MLP encoder $\xi$ to generate output representation $H_f$ of SAGN:
\begin{equation}
    H_f=\xi(H_{att} + XW_r),
\end{equation}
where $W_r$ is the residual linear matrix.


\subsection{SLE}

To incorporate label information into scalable GNNs, we remove inner random masking within the training set to process label information in a more general way. We add the base model with a scalable label model with the lastly smoothed label vectors as input. The self-training approach can be directly implemented on scalable GNNs. We further allow hard pseudo labels at each stage to participate label propagation with true labels. The hard pseudo labels with high confidence brings model knowledge into label propagation.

\paragraph{Self-training process}
We split training process into stages. At the first stage, we train a model with the raw training set. Starting from the second stage, we enhance the training set with the previously predicted probabilities and train a new model. After certain stages we obtain the final model.

To enhance the training set, we firstly obtain the confident node set at the $s$-stage $\overline{\mathcal L}_s$ ($s>0$) with a threshold $\beta$ as below:
\begin{equation}
    \overline{\mathcal L}_s=\{i,\mathop{\rm max}_c(\hat{Y}_{s-1,i,c})\geq \beta\}\subset \mathcal N.
\end{equation}
Then the enhanced training set at the $s$-stage $\mathcal L_s$ is updated as below:
\begin{equation}
    \mathcal L_s=\mathcal L_0 \cup \overline{\mathcal L}_s.
\end{equation}

\begin{figure}[tb]
    \centering
    \includegraphics[width=0.50\linewidth]{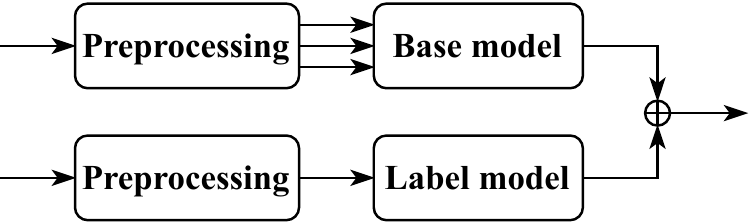}
    \caption{Combination of base model and label model in SLE. The node feature information and label information are respectively passed through the base model (SAGN) and label model.}
    \label{fig:SLE}
\end{figure}

\begin{figure}[tb]
    \centering
    \includegraphics[width=0.50\linewidth]{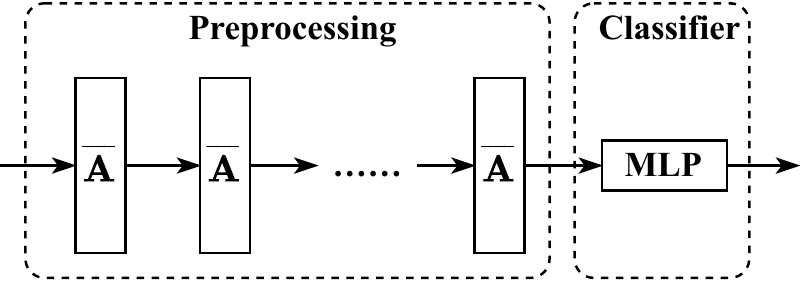}
    \caption{Architecture of the label model.}
    \label{fig: label model}
\end{figure}

\paragraph{Label propagation at each stage}
At each stage, we add the base model with the label model, as shown in Figure \ref{fig:SLE}. We perform label propagation to generate input of the label model, as shown in Figure \ref{fig: label model}.

Firstly, the initial 0-hop label embedding is generated. We set the initial label embedding at the first stage $\overline{Y}^{(0)}_0=Y$. From the second stage, the initial label embedding $\overline{Y}^{(0)}_{s}$ is obtained with ground truth labels and hard pseudo labels:
\begin{equation}
\label{eq: update initial label embedding}
\begin{split}
    \overline{Y}^{(0)}_{s,i}=\begin{cases} 
                                Y_{i},i\in \mathcal L_0 \\
                                \tilde{Y}_{s-1,i}, i \in \mathcal L_{s}\setminus \mathcal L_0\\
                                \mathbf{0}, i \in \mathcal U_{s}
                             \end{cases},
\end{split}
\end{equation}
where $i$ represents the node $i$ and $\mathbf{0}$ represents $C$-dimension zero vector.

Secondly, the label propagation with a maximum hop of $K_l$ is computed and the label model $\phi_s$ is applied to the last label embedding $\overline{Y}^{(K_l)}_s=\overline{A}^{K_l}\overline {Y}^{(0)}_s$ as below:
\begin{equation}
\label{eq: update label embedding}
(H_l)_s=\phi_s(\overline{A}^{K_l}\overline {Y}^{(0)}_s).
\end{equation}


Finally, with using SAGN as the base model for example, we sum outputs from the base model (with adding the subscript $s$) and label model to obtain the final representation matrix $H_s$ as below:
\begin{equation}
H_s=
 \xi_s\left(\sum_{k=0}^{K_f}\Theta^{(k)}_s\zeta^{(k)}_s\left(\overline{A}^k X\right) + X(W_r)_s\right)  + \phi_s\left(\overline{A}^{K_l} \overline{Y}^{(0)}_{s}\right).
\end{equation}



\paragraph{Loss function and optimization}
We do not use Knowledge Distillation (KD) loss \cite{hinton2015distilling}, since it conflicts with hard pseudo labels and empirically brings no improvements. The loss function of SLE at the $s$-stage is calculated as below:

\begin{equation}
L_s = -\frac{1}{|\mathcal L_s|}\left(\sum_{i}\sum_{c}{Y_{i,c}\mathop{\rm log}(\hat Y_{s,i,c})}+\sum_{j}\sum_{c}{\tilde{Y}_{s-1,j,c}\mathop{\rm log}(\hat Y_{s,j,c})}\right),
\end{equation}
where $c$ represents the $c$-th class, $|\mathcal L_s|$ is the element number of $\mathcal L_s$, $i$ satisfies $i \in \mathcal L_0$ and $j$ satisfies $j \in \mathcal L_{s} \setminus \mathcal L_0$. Based on this loss function, we update model parameters with the Adam optimizer \cite{kingma2014adam}.

\begin{table}[tb]
\caption{Summary of datasets. For classes, there are single (s) label classification task and multiple (m) label classification task.}
\label{tab: summary of datasets}
\begin{center}
    \setlength{\tabcolsep}{1mm}{
    \begin{tabular}{lccccc}
    \toprule
       Dataset & Nodes & Edges & Classes & Train/Val/Test & Setting\\
       \midrule
       Reddit & 232,965 & 11,606,919 & 41(s) & 66\% / 10\% / 24\% & Inductive \\
       Yelp & 716,847 & 6,977,410 & 100(m) & 75\% / 10\% / 15\% & Inductive \\
       Flickr & 89,250 & 899,756 & 7(s) & 50\% / 25\% / 25\% & Inductive\\
       PPI & 14,755 & 225,270 & 121(m) & 66\% / 12\% / 22\% & Inductive \\
       ogbn-products & 2,449,029 & 61,859,140 & 47(s) & 10\% / 2\% / 88\% & Transductive\\
       ogbn-papers100M & 111,059,956 & 1,615,685,872 & 172(s) & 78\% / 8\% / 14\% & Transductive\\
       ogbn-mag & 1,939,743	& 21,111,007 & 349(s) & 85\% / 9\% / 6\% & Transductive \\
    \bottomrule
    \end{tabular}
    }
\end{center}

\end{table}

\section{Experiments and analysis}
\paragraph{Datasets}
We evaluate our proposed method oh both inductive (Reddit, Flickr, PPI and Yelp) \cite{hamilton2017inductive, zeng2019graphsaint} and transductive (ogbn-products, ogbn-papers100M and ogbn-mag) \cite{hu2020open} datasets. The detailed descriptions of datasets and experiments on heterogeneous ogbn-mag are reported in Appendices. The statistics of these datasets are listed in Table \ref{tab: summary of datasets}.

\paragraph{Inductive settings}
In the inductive setting \cite{hamilton2017inductive, zeng2019graphsaint, zitnik2017predicting}, the nodes out of the training set are forbidden to participate in model training but allowed in inference. With SLE, from the second stage, we actually use node features out of the raw training set when training. However, we do not rely on the prior that the nodes from different sets have inter-set edges in transductive setting. Moreover, we still restrict graph convolutions in raw train graph(s) for nodes belonging to the raw training set when training with the enhanced training set. It meets the actual condition when applying the model on unseen data. 

\paragraph{Special notations}
In experimental results, $s$-SLE ($s>0$) means that we train the model under SLE approach after $s$ stages. SE refers to SLE approach without the label model, which is equal to the self-training approach\cite{li2018deeper}. 0-SLE means that we train the base model with the label model using the raw training set. The notation 0-SE is negligible since there are no modifications.

\paragraph{Baselines and hardware configuration}
We select vanilla GCN \cite{kipf2016semi,schlichtkrull2017modeling}, UniMP \cite{shi2020masked}, sampling-based GNNs \cite{hamilton2017inductive, ying2018graph, huang2018adaptive, chen2018fastgcn, chen2017stochastic, zou2019layer, chiang2019cluster, zeng2019graphsaint} and sampling-free methods \cite{wu2019simplifying, rossi2020sign, huang2020combining, grover2016node2vec, dong2017metapath2vec, yu2020scalable, bordes2013translating} as baselines. For baselines we use metrics from official reports and Open Graph Benchmark (OGB) leaderboard \cite{hu2020open}. For our proposed methods, we conduct experiments with Pytorch (with a BSD-style license) \cite{paszke2019pytorch} and Deep Graph Library (DGL, with Apache License 2.0 license) \cite{wang2019dgl} on a China Telecom Cloud p2v.16xlarge.8 instance, with 512G memory, 8 NVIDIA V100 GPUs, 64 vCPUs and a processor Intel(R) Xeon(R) Gold 6151 CPU @ 3.00GHz. We run for 10 times with the same random seed setting (from 0 to 9) for each experiment.

\paragraph{Setup}
The transition matrix is selected between symmetrically normalized adjacency matrix and row-stochastic random walk matrix. 
The MLP encoders in SAGN share the same architecture with the same linear layer number, batch normalizations and ReLU activation functions. The label model doubles the layer number of SAGN. For multi-class tasks, we simply filter confident nodes whose maximum probability is larger than a certain threshold (between 0 and 1). For multi-label tasks, we filter confident nodes whose mean entropy is smaller than a computed threshold between 0 and $\rm log2$. Moreover, for multi-label tasks, we set the initial label embedding of nodes out of the training set to 0.5. Besides of standard weight decay and dropout techniques, we use attention dropout and input dropout.

\begin{table}[t]
\caption{Results on inductive datasets. The means and standard deviations of micro-F1 score over 10 runs are reported. The best results are highlighted in \textbf{bold} fonts.}\label{table: inductive}
\begin{center}
\begin{tabular}{cccccc}
\toprule
Method & Reddit & Flickr &  PPI & Yelp \\
\midrule
GCN \cite{kipf2016semi}            & 93.3±0.0\% & 49.2±0.3\% & 51.5±0.6\% & 37.8±0.1\% \\
FastGCN \cite{chen2018fastgcn}     & 92.4±0.1\% & 50.4±0.1\% & 51.3±3.2\% & 26.5±5.3\% \\
Stochastic-GCN \cite{chen2017stochastic}  & 96.4±0.1\% & 48.2±0.3\% & 96.3±1.0\% & 64.0±0.2\%\\
AS-GCN \cite{huang2018adaptive}     & 95.8±0.1\% & 50.4±0.2\% & 68.7±1.2\% & —\\
GraphSAGE \cite{hamilton2017inductive}      & 95.3±0.1\% & 50.1±1.3\% & 63.7±0.6\% & 63.4±0.6\%\\
ClusterGCN \cite{chiang2019cluster}     & 95.4±0.1\% & 48.1±0.5\% & 87.5±0.4\% & 60.9±0.5\% \\
GraphSaint \cite{zeng2019graphsaint}     & 96.6±0.1\% & 51.1±0.1\% & \textbf{98.1±0.4\%} & \textbf{65.3±0.3\%}\\
SGC \cite{wu2019simplifying}& 94.9±0.0\% & 50.2±0.1\% & 89.2±1.5\% & 35.8±0.6\%\\
SIGN \cite{rossi2020sign}            & 96.8±0.0\% & 51.4±0.1\% & 97.0±0.3\% & 63.1±0.3\%\\
SAGN          & 96.9±0.0\% & 51.4±1.2\% & 97.9±0.1\% & \textbf{65.3±0.1\%}\\ 
SAGN+1-SLE    & \textbf{97.1±0.0\%} & 54.3±0.5\% & 98.0±0.1\% & \textbf{65.3±0.1\%}\\
SAGN+2-SLE    & \textbf{97.1±0.0\%} & \textbf{54.6±0.4\%} & 98.0±0.1\% & \textbf{65.3±0.1\%}\\
\bottomrule
\end{tabular}
\end{center}
\end{table}

\paragraph{Inductive results}
For inductive datasets, at the first stage of SLE, we do not use the label model since the label propagation is restricted in the training set. As shown in Table \ref{table: inductive}, SAGN outperforms baselines in most inductive datasets. For Reddit and Flickr, SAGN achieves close results to SIGN. For PPI and Yelp, SAGN outperforms SIGN by respectively 0.9\% and 2.2\%. It implies that the attention mechanism in SAGN is more effective for some inductive datasets. SAGN+1-SLE and SAGN+2-SLE both outperform their previous models. Although for most inductive datasets SLE brings few improvements, except that SLE improves performance significantly by 3.2\% in Flickr. We can attribute to that in Flickr the label distribution follows smoothness assumption. For Yelp we remove the label model at all stages because empirically it brings no improvements. We hypothesise that in Yelp the smoothness assumption of labels is not well satisfied. Note that all hyperparameters are hand-tuned, while hyperparameters of SIGN are tuned using Bayesian optimization \cite{bergstra2011algorithms} on inductive datasets.


\begin{table}[t]
\caption{Results on transductive datasets. The means and standard deviations of validation and test accuracies over 10 runs are reported. The best results are highlighted in \textbf{bold} fonts.}\label{ogbn-products and ogbn-papers100M}
\begin{center}
\begin{tabular}{ccccc}
\toprule
\multirow{3}*{Method} & \multicolumn{2}{c}{ogbn-products} & \multicolumn{2}{c}{ogbn-papers100M} \\
\cmidrule(r){2-3} \cmidrule(r){4-5}
 & Validation & Test & Validation & Test\\
\midrule
MLP & 75.54±0.14\% & 61.06±0.08\% & 49.60±0.29\% & 47.24±0.31\%\\
Node2Vec \cite{grover2016node2vec, hu2020open} & 90.32±0.06\% & 72.49±0.10\% & 58.07±0.28\% & 55.60±0.23\% \\
GCN \cite{kipf2016semi, hu2020open} & 92.00±0.03\% & 75.64±0.21\% & — & — \\
GraphSAGE \cite{hamilton2017inductive, hu2020open} & 92.24±0.07\% & 78.50±0.14\% & — & — \\
NeighborSampling \cite{hamilton2017inductive, hu2020open}& 91.70±0.09\% & 78.70±0.36\% & — & — \\
ClusterGCN \cite{chiang2019cluster} & 92.12±0.09\% & 78.97±0.33\% & — & — \\
GraphSaint \cite{zeng2019graphsaint}& — & 80.27±0.26\% & — & —\\
SIGN \cite{rossi2020sign, wang2019dgl}& 92.86±0.02\% & 80.52±0.13\% &69.32±0.06\% & 65.68±0.06\% \\
SAGN & 93.09±0.04\% & 81.20±0.07\% & 70.34±0.99\% & 66.75±0.84\% \\
SAGN+1-SE & 92.54±0.04\% & 82.23±0.09\% & 70.79±0.12\% & 67.21±0.12\% \\
SAGN+2-SE & 92.33±0.03\% & 82.50±0.13\% & 70.89±0.12\% & 67.30±0.15\% \\
\midrule
UniMP & 93.08±0.17\% & 82.56±0.31\% & 71.72±0.05\%	 & 67.36±0.10\% \\
MLP+C\&S & 91.47±0.09\% & 84.18±0.07\% & — & — \\
SAGN+0-SLE &  93.27±0.04\%  &  83.29±0.18\% & 71.06±0.08\%  &  67.55±0.15\% \\
SAGN+1-SLE &  93.06±0.07\%  &  84.18±0.14\% & 71.23±0.10\%  &  67.77±0.15\% \\
SAGN+2-SLE &  92.87±0.03\%  &  \textbf{84.28±0.14\%} & 71.31±0.10\%  &  \textbf{68.00±0.15\%} \\
\bottomrule
\end{tabular}
\end{center}
\end{table}

\paragraph{Transductive results}
For ogbn-products, as shown in Table \ref{ogbn-products and ogbn-papers100M}, SAGN outperforms SIGN by 0.68\%. And SIGN outperforms MLP, Node2Vec, vanilla GCN and sampling-based methods by at least 0.25\%. This implies that the inception-like module, MLP encoders and attention mechanism can bring stronger expressiveness. SAGN can be further improved by up to 1.30\% with self-training. For methods using labels, SAGN+2-SLE achieves state-of-the-art results with 3.08\% improvements based on SAGN. The performance also improves with stages increasing. Because SLE can tap potential of the model with using label information. MLP+C\&S also achieves close results, but it requires complex (though partial) decomposition of laplacian matrix.
For the largest dataset ogbn-papers100M, as shown in Table \ref{ogbn-products and ogbn-papers100M}, SAGN outperforms other methods without using label information by at least 1.07\%. It can also be improved by 0.55\% with self-training. With improvements of 1.25\% based on SAGN, SAGN+2-SLE achieves state-of-the-art results. Full batch GCN, C\&S and sampling-based methods have not been implemented for this dataset. UniMP is implemented with massive GPU memory. It is still outperformed by SAGN+SLE even at the first stage.


\begin{table}[t]
\caption{Time complexity analysis for scalable methods.}\label{tab: complexity}
\begin{center}
    \setlength{\tabcolsep}{1.2mm}{
    \begin{tabular}{cccc}
    \toprule
         Method& Multi-hop encoders & Post encoder & Total  \\
    \midrule
         SIGN & $\mathcal{O}(KLNd^2)$ & $\mathcal{O}((K+L)Nd^2)$ & $\mathcal{O}(KLNd^2)$\\
         SAGN & $\mathcal{O}(KLNd^2)$ & $\mathcal{O}((K/d+L)Nd^2)$ & $\mathcal{O}(KLNd^2)$\\
         MLP  & — & $\mathcal{O}(LNd^2)$ & $\mathcal{O}(LNd^2)$ \\
    \bottomrule
    \end{tabular}
    }
\end{center}
\end{table}

\paragraph{Complexity}
The complexity of SAGN is similar to SIGN \cite{rossi2020sign}. Because they share the same preprocessing and similar architecture in post classifier. The complexity from the attention mechanism is negligible compared to multi-hop encoders. We suppose that the layer numbers of multi-hop MLP encoders and post MLP encoder share the same order of magnitude $L$. We use $d$ to represent hidden dimension and $K$ to represent number of hops.  
The time complexity of SIGN and SAGN are shown in Table \ref{tab: complexity}. SAGN has less complexity in post encoder. Finally, they share the same total time complexity $KLNd^2$ which is $K$ times larger than MLP with the same layer number.

\begin{table}[tb]
\caption{Runtime experiments on ogbn-products. The runtimes (seconds) with means and standard deviations, memory costs (Mb) and parameter numbers over 10 runs are reported. SIGN+SLE is not reported due to out-of-memory (OOM) error.}\label{tab: runtime experiments}
\begin{center}
    \begin{tabular}{ccccc}
    \toprule
       Method & Training & Inference & Memory & Parameters\\
    \midrule
       MLP          & 0.67±0.04s  & 7.78±0.25s & 10832Mb & 669743\\
       SIGN         & 1.27±0.10s   & 8.68±0.64s & 15558Mb & 3489847\\
       SAGN         & 1.12±0.01s  & 8.43±0.22s & 13948Mb & 2233391\\
       \midrule
       MLP+0-SLE  & 0.76±0.03s   & 7.85±0.30s & 11958Mb & 1181278\\
       SAGN+0-SLE & 1.47±0.01s        & 8.35±0.33s & 14894Mb & 2810462\\

    \bottomrule
    \end{tabular}
\end{center}
\end{table}

\paragraph{Runtime}
The runtime comparison between SIGN and sampling-based methods has been previously conducted \cite{rossi2020sign}. In conclusion, SIGN has longer preprocessing runtime but faster training runtime and much faster inference runtime. We conduct runtime experiments on dataset ogbn-products with scalable baselines including MLP and SIGN. As shown in Table \ref{tab: runtime experiments}, SIGN and SAGN have very close runtime in both training and inference process, as expected in complexity analysis. SAGN has 10\% smaller memory cost and 36\% less parameters compared with SIGN, which comes from the removal of concatenation operation. The label model brings longer training time, a bit more memory cost and more parameters. However, the inference runtime seems stable for each base model, since there is no gradient computation in inference.

\begin{table}[tb]
\caption{Ablation studies on ogbn-products. The means and standard deviations of validation and test accuracies over 10 runs are reported. SIGN+SLE is not reported due to OOM error. The best results in every stage are highlighted in \textbf{bold} fonts.}
    \label{tab:ablation studies}
\begin{center}

    \begin{tabular}{cccc}
    \toprule
       Method & 0-stage & 1-stage & 2-stage \\
       \midrule
       MLP+SE & 75.75±0.12\% & 75.11±0.18\%& 74.99±0.19\%\\
       SIGN+SE & 80.52±0.13\% & 81.42±0.14\% & 81.56±0.17\% \\
       SAGN*+SE & 80.25±0.09\% & 81.35±0.08\% & 81.47±0.11\%\\
       SAGN**+SE & 80.04±0.07\% & 81.16±0.10\% & 81.29±0.11\%\\
       SAGN+SE  & 81.20±0.07\% & 82.23±0.09\% & 82.50±0.13\%\\
       \midrule
       SLE  & 78.86±1.48\% & 77.47±0.55\% & 77.58±0.89\%\\
       MLP+SLE & 82.42±0.48\% & 83.76±0.26\% & 83.91±0.26\% \\
       SAGN*+SLE & 82.49±0.27\% & 83.50±0.32\% & 83.56±0.30\%\\
       SAGN**+SLE & 82.52±0.27\% & 83.56±0.27\% & 83.63±0.25\% \\
       SAGN+SLE & \textbf{83.29±0.18\%} & \textbf{84.18±0.14\%} & \textbf{84.28±0.14\%}\\
   \bottomrule
    \end{tabular}

\end{center}
\end{table}

\paragraph{Ablation studies}
The ablation studies are conducted on ogbn-products. SAGN* and SAGN**, based on SAGN, respectively replace attention weights with uniform and exponentially decaying (with ratio of 0.5) weights. MLP uses the input of diffused feature matrix \cite{klicpera2019diffusion}. As shown in Table \ref{tab:ablation studies}, both SAGN* and SAGN** are outperformed by SAGN with SE and SLE, which proves the effectiveness of the attention mechanism. The results of SAGN* and SIGN are very close as expected. Each model with SLE outperforms the same base model with SE by at least 1.81\%, which proves the effectiveness of the label model in SLE. Interestingly, MLP+SLE obtains better results than SAGN*+SLE and SAGN**+SLE. But MLP has worse results and cannot be improved with self-training. On the other hand, single label model (SLE in Table \ref{tab:ablation studies}) cannot either be improved with self-training. The combination of them brings instant improvement of at least 3.56\%. Moreover, the combined model can be further improved by up to 1.49\% with the enhanced training set. We hypothesise that MLP captures less overlapping information with the label model than other base models. These results highlight the effectiveness of combination of the base model and label model, which brings both instant and potential improvements.

\section{Conclusion}
We propose sampling-free SAGN and training approach SLE for semi-supervised graph learning on large scaled data. With similar complexity and runtime to SIGN, SAGN achieves better expressiveness. SLE can further boost many base models including MLP, SIGN and SAGN. SAGN+SLE outperforms both sampling-based and sampling-free baselines on most datasets. Our proposed methods are easy to implement and extend on huge graphs with competitive performances. 

\paragraph{Simplicity and expressiveness}
Scalable methods actually make a tradeoff between simplicity and expressiveness \cite{rossi2020sign}. The precomputed graph convolutions may lose some initial information. To ensure considerable performance, it is intuitive to design complex post classifier with graph structure-aware components. For examples, SIGN using MLP and inception-like module outperforms SGC and SAGN using the adaptive attention mechanism outperforms SIGN.

\paragraph{Limitations}
The transition matrix can play a key role in performance \cite{rossi2020sign}. However, we focus on modifying post classifier to obtain better expressiveness with using only two kinds of transition matrices in this paper. 
The multi-hop input requires large CPU memory cost proportional to hop number. 
SLE in some inductive datasets brings less improvements.
The enhanced training set with more nodes bring longer training runtime but faster converge speed. We can cutoff training at earlier steps.

\bibliographystyle{unsrt}
\bibliography{neurips_2021}

\newpage







\begin{appendix}
\section{Details of models}
\subsection{Concatenation operation in SIGN}
The concatenation operation with linear layer followed is equal to applying multiple linear layers and sum them up. This conclusion can be directly obtained by rules of matrix multiplication:
\begin{equation}
    \left(\Big |\Big |_{k=0}^{K}H^{(k)}\right)W = \sum_{k=0}^{K}\left(H^{(k)} W^{(k)}\right).
\end{equation}

$W^{(k)}$ is the $k$-th row slice of $W$ with length of $d$:

\begin{equation}
    W^{(k)} = W_{[kd:(k+1)d,:]},
\end{equation}
where $d$ is the hidden dimension of $H^{(k)}$. 

There are no tunable or learnable weights in the concatenation operation. It also increases the memory cost. Thus we explicitly sum representation matrices with learnable attention weights, which improves expressiveness and reduces memory cost with considerable additional complexity.

\subsection{SLE}
The overall process of SLE approach is presented in Algorithm \ref{algo: SLE}. The graph convolutions of node features are computed at once. During each stage after 0-stage, the model is trained with enhanced train set including more train nodes ($\mathcal L_s \setminus \mathcal L_0$), updated pseudo hard labels ($\tilde Y_{s-1}$) and label embeddings ($\overline{Y}^{(K_f)}_s$). With hard pseudo labels involved, the label propagation (or graph convolutions of label embeddings) should be recomputed with updated initial 0-hop label embedding. 

\begin{algorithm}[htb]
  \SetAlgoLined
  \KwData{transition matrix $\overline{A}$, node feature $X$, node one-hot encoded labels $Y$ (zeros on validation and test sets), node set $\mathcal N$, class number $C$, set of labeled nodes (train set) $\mathcal L_0$, set of unlabeled nodes $\mathcal U_0$, maximum feature convolution hop $K_f$, maximum label propagation hop $K_f$, maximum stage $S$, threshold $\beta$}
  \KwResult{final model $f_S+g_S$}
  $X^{(0)}=X$\;
  $Y^{(0)}_0=Y$\;
  $k=1$\;
  \While{$k\leq K_f$}{
      $\overline {X}^{(k)}=\overline{A}\overline {X}^{(k-1)}$\;
      $k=k+1$\;}
  $k=1$\;
  \While{$k\leq K_l$}{
      $\overline {Y}^{(k)}_{0}=\overline{A}\overline {Y}^{(k-1)}_{0}$\;
      $k=k+1$\;}
      
  train the first model $f_0+g_0$ with $\mathcal L_0$ and $\mathcal U_0$\; 
  output soft labels $\hat{Y}_0$ and hard labels $\tilde{Y}_0$\;
  $s=1$\;
  \While{$s \leq S$}{
    $\overline{\mathcal L}_s=\{i, \mathop{\rm max}_c(\hat{Y}_{s-1,i,c})\geq \beta\}\subset \mathcal N$\;
    $\mathcal L_s=\mathcal L_0\cup \overline{\mathcal L}_s$, $\mathcal U_s=\mathcal N\setminus \mathcal L_s$\;
    
    $\overline{Y}^{(0)}_{s,i}=\begin{cases} 
                                Y_{i},i\in \mathcal L_0 \\
                                \tilde{Y}_{s-1,i}, i \in \mathcal L_{s}\setminus \mathcal L_0\\
                                [0, 0, ..., 0], i \in \mathcal U_{s}
                             \end{cases}$\;

    $k=1$\;
      \While{$k\leq K_l$}{
          $\overline {Y}^{(k)}_{s}=\overline{A}\overline {Y}^{(k-1)}_{s}$\;
          $k=k+1$\;}
    train the $s$-th model $f_s+g_s$ using $\mathcal L_s$ and $\mathcal U_s$\;
    output soft labels $\hat{Y}_s$ and hard labels $\tilde{Y}_s$\;
    $s=s+1$\;
    }
  \caption{Self-Label-Enhanced training approach}
  \label{algo: SLE}
\end{algorithm}

\subsection{Hard pseudo labels}
We aim to recover the label distribution of unlabeled nodes and propagate them with true labels in graphs. Unlike Knowledge Distillation, the recovered labels should be hard labels to align with the existed true labels. From another aspect, the hard pseudo labels without detailed probability distributions incorporates some moderate "noises" into teacher model's knowledge. With label propagation, these "noises" can also be propagated into other nodes. Then the model based on enhanced train set can more easily capture new knowledge. Meanwhile, to get high-quality pseudo labels and limit noises, we filter "confident" nodes. For examples, we can use a threshold-based or top-K filter to get confident nodes.

\section{Details of datasets and settings}


\subsection{Dataset details}

Reddit \cite{hamilton2017inductive} contains nodes representing posts in Reddit website with 300-dimensional GloVe CommonCrawl word vectors. An edge represents that two posts are commented by the same user. The node label represents its community. This is an inductive multi-class classification task. The time range of dataset is during the month of September, 2014. The first 20 days are used for training and the remaining days for testing.

In Flickr \cite{zeng2019graphsaint}, the nodes represent pictures uploaded to the Flickr website with links representing they are from the same location, submitted to the same gallery, sharing common tags, etc. The node feature contains information of low-level feature. It is aimed to predict the tag of pictures, which is an inductive multi-class classification task. The train, validation and test nodes are randomly split into 0.5, 0.25 and 0.25.

PPI \cite{hamilton2017inductive} contains multiple subgraphs representing human tissues with proteins as nodes and protein interactions as edges. The positional gene sets, motif gene sets and immunological signatures are used as node features. We aim to predict protein roles, which is an inductive multi-label classification task. 20 graphs are used for training. Two graphs are used for testing with another two graphs for validation.

In Yelp \cite{zeng2019graphsaint}, each node represents an active user and each edge represents a friendship relation. The node feature is generated from reviews of users using Word2Vec model. The task is to predict the types of business that the user has been to, which is an inductive multi-label classification task. The train, validation and test nodes are randomly split into 0.75, 0.10 and 0.15.

For ogbn-products \cite{hu2020open}, the graph represents an Amazon product co-purchasing network. Nodes represent products sold in Amazon. Edges represent that prodcuts are purchased together. Node features are 100-dimensional bag-of-words features from their descriptions. The task is to predict the category of a product, which is a transductive multi-class classification task. Instead of randomly splitting, The sales ranking (popularity) is used to split nodes into training, validation and test sets. Firstly the products is sorted according to their sales ranking. Secondly the top 8\% is used for training, next top 2\% is used for validation, and the rest is used for testing.

ogbn-papers100M \cite{hu2020open} is the largest node classification dataset. It contains up to 111 million nodes representing papers indexed by MAG and 1.6 billion edges representing citations. Each node has a 128-dimension feature vector by averaging the word embeddings of its title and abstract. The task is to predict the subject ares of the subset of papers published in arXiv (about 1.5 million). This is a transductive multi-class classification task. The training nodes with labels are all arXiv papers published until 2017, while the validation nodes are the arXiv papers published in 2018, and the models are tested on arXiv papers published since 2019.

ogbn-mag \cite{hu2020open} is a heterogeneous graph which is a subset from MAG. There are 736,389 "paper" nodes, 1,134,649 "author" nodes, 8,740 "institution" nodes and 59,965 "fields of study" nodes. For relations, an author "writes" a paper, an author is "affiliated with" an institution, a paper "cites" another paper and a paper "has a topic of" a field of study. Each paper node has a 128-dimension Word2Vec feature vector and other types of nodes do not have feature vectors. We aim to predict the venue of papers, which is a multi-class classification task. The way of splitting train, validation and test sets is the same as ogbn-papers100M.

All datasets are open sourced with MIT license.

\subsection{Inductive setting}
For most of datasets, all nodes are in the same graph. In PPI, the train, validation and test sets are totally separated without any inter-set edges. The propagation of both node feature and node label information is restricted in subgraphs. Although it is possible to apply SLE in such completely inductive setting, the distributions of label information vary between train set and other sets. Because we have full true labels in train sets but only parts of validation and test sets are filtered with pseudo labels. Even after smoothing, the label embeddings in train set and other sets can be very different. To further balance the distributions of label embedding in different sets, we also filter confident nodes in raw train set when generating initial label embedding. It means that, in raw train set, only nodes with high confidence have associated non-zero initial label embedding. In addtition, the other nodes with zero initial label embedding still participate in model training.

\subsection{Baseline selection}
For Reddit, Flickr, PPI and Yelp, we select vanilla GCN \cite{kipf2016semi}, sampling-based GNNs \cite{hamilton2017inductive, ying2018graph, huang2018adaptive, chen2018fastgcn, chen2017stochastic, zou2019layer, chiang2019cluster, zeng2019graphsaint} and sampling-free GNNs \cite{wu2019simplifying, rossi2020sign} as baselines. The neighbor sampling based GNNs include FastGCN, Stochastic-GCN, AS-GCN, GraphSAGE, CluterGCN and GraphSaint. The sampling-free GNNs include SGC and SIGN.

For ogbn-products, MLP \cite{hu2020open}, Node2Vec \cite{grover2016node2vec}, full-batch GCN \cite{kipf2016semi}, full-batch GraphSAGE \cite{hamilton2017inductive}, sampling-based GNNs \cite{hamilton2017inductive, hu2020open, chiang2019cluster, zeng2019graphsaint, shi2020masked} and sampling-free methods \cite{rossi2020sign, huang2020combining} are used as baselines. The sampling-based GNNs include NeighborSampling (SAGE), ClusterGCN, GraphSaint and UniMP. The scalable methods include MLP, SIGN and MLP+C\&S. UniMP, MLP+C\&S and SAGN+SLE incorporate label information.

For ogbn-papers100M, MLP \cite{hu2020open}, Node2Vec \cite{grover2016node2vec}, UniMP \cite{shi2020masked} and SIGN \cite{rossi2020sign} are used as baselines. UniMP and SAGN+SLE incorporate label information.

For ogbn-mag, MLP \cite{hu2020open}, vanilla GCN \cite{kipf2016semi}, GraphSAGE \cite{hamilton2017inductive} and SIGN \cite{rossi2020sign} are used as baselines in homogeneous setting. MetaPath2Vec \cite{dong2017metapath2vec}, vanilla full-batch R-GCN \cite{schlichtkrull2017modeling}, NeighborSampling (R-GCN) \cite{hu2020open}, ClusterGCN with R-GCN aggregator \cite{chiang2019cluster, schlichtkrull2017modeling, hu2020open}, GraphSaint with R-GCN aggregator \cite{zeng2019graphsaint, chiang2019cluster, hu2020open} and NARS \cite{yu2020scalable} are used as baselines in heterogeneous setting.

\subsection{Setup}
Specially, for ogbn-mag, we convert heterogeneous graph into homogeneous graph. However, nodes other than paper nodes have no raw node features. We follow two types of solutions: 1. Average their neighboring paper nodes' attributes to obtain equivalent attributes. 2. Use embedding vectors from pretrained TransE model. The latter also incorporates heterogeneous information in pretrained embeddings. The TransE embedding has been proven to be useful for heterogeneous graphs.

\section{Other experiments}
\subsection{Experiments on ogbn-mag}
\begin{table}[htb]
\caption{Results on ogbn-mag. Validation and test accuracies with means and standard deviations (\%) are reported. The best results in homogeneous and heterogeneous settings are highlighted in \textbf{bold} fonts.}\label{ogbn-mag}
\begin{center}
\begin{tabular}{ccc}
\toprule
Method & Validation & Test\\
\midrule
MLP & 26.26±0.16\% & 26.92±0.26\%\\
GCN & 29.53±0.22\% & 30.43±0.25\%\\
GraphSAGE & 30.70±0.19\% & 31.53±0.15\% \\
SIGN & 40.68±0.10\% & 40.46±0.12\% \\
SAGN+0-SLE & 42.13±0.51\% & 40.51±0.83\%\\
SAGN+1-SLE & 43.85±0.49\% & 42.18±0.61\%\\
SAGN+2-SLE & 44.27±0.30\% & \textbf{42.75±0.38\%}\\
\midrule
MetaPath2Vec & 35.06±0.17\% & 35.44±0.36\%\\
R-GCN & 40.84±0.41\% & 39.77±0.46\%\\
NeighborSampling & 47.61±0.68\% & 46.78±0.67\%\\
ClusterGCN & 38.40±0.31\% & 37.32±0.37\% \\
GraphSaint & 48.37±0.26\% & 47.51±0.22\% \\
NARS & 53.72±0.09\% & 52.40±0.16\% \\
SAGN+TransE+0-SLE & 49.42±0.17\% & 47.95±0.25\%\\
SAGN+TransE+1-SLE & 51.23±0.16\% & 49.70±0.17\%\\
SAGN+TransE+2-SLE & 51.80±0.15\% & 50.29±0.16\%\\
NARS\_SAGN+0-SLE & 54.12±0.15\% & 52.32±0.25\%\\
NARS\_SAGN+1-SLE & 55.52±0.16\% & 53.95±0.14\%\\
NARS\_SAGN+2-SLE & 55.91±0.17\% & \textbf{54.40±0.15\%}\\
\bottomrule
\end{tabular}
\end{center}
\end{table}

\begin{figure}
    \centering
    \includegraphics[width=0.50\linewidth]{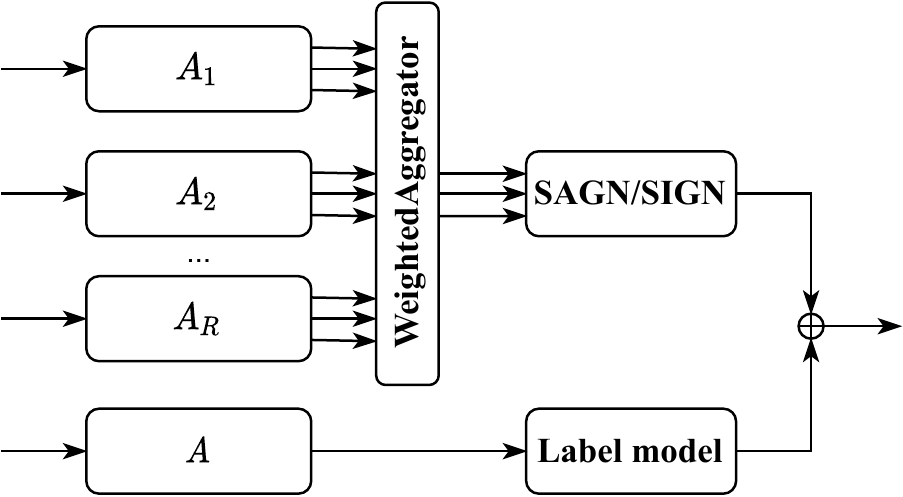}
    \caption{NARS+SLE. $A_r$ refers to the subgraph induced by the $r$-th relation subset. $A$ refers to the whole graph. Each subgraph can generate $(K+1)$ feature matrices of different hops. The weighted aggregator is used to combine feature matrices from different subgraphs. Then the combined $(K+1)$ feature matrices are forwarded into SAGN or SIGN. By adding output of the node label model, NARS can be used as the base model in SLE.}
    \label{fig: NARS}
\end{figure}

Table \ref{ogbn-mag} shows the results on ogbn-mag dataset with both homogeneous and heterogeneous settings. The means and standard deviations of validation and test accuracies are reported. For homogeneous settings, naive MLP acheives low test accuracy with mean 26.92\%. Vanilla GCN improves the mean test accuracy to 30.43\%. Sampling-based GraphSAGE improves GCN's result by 1\%. Scalable SIGN effectively improves the mean test accuracy up to 40.46\%. SAGN at the 1-stage slightly outperforms SIGN. But with enhanced train set SAGN brings the metric up to 42.75\%. For heterogeneous setting, which more fits the heterogeneous graphs, MetaPath2Vec acheives 35.44\% mean test accuracy. The vanilla full-batch R-GCN is outperformed by sampling-based methods except ClusterGCN. The cluster partitioning in ClusterGCN may in somehow break the initial data distribution. We use TransE embedding (with the dimension of 128) to incorporate heterogeneous information into SAGN. At the 0-stage, SAGN outperforms other naive and sampling-based methods. With stage increasing, SAGN improves the metric up to 50.29\%. This score is higher than most of GNNs specially designed for heterogeneous graphs except NARS. NARS \cite{yu2020scalable} uses TransE \cite{bordes2013translating} embedding (with higher dimension of 256) and subgraphs induced by sampled relation subsets, which is also based on SIGN. By experiments on heterogeneous ogbn-mag dataset with directly adding TransE embedding, we prove that our proposed SAGN with SLE can achieve better or comparable results compared with other heterogeneous GNNs. We believe that incorporating heterogeneous model architecture into SAGN will bring better performance in heterogeneous graph learning tasks. Thus we simply replace SIGN component in NARS with SAGN to obtain NARS\_SAGN model. The detailed model architecture is shown in Figure \ref{fig: NARS}. We use 256-dimension TransE embedding this time. The example relation subsets from NARS are used. NARS\_SAGN+SLE achieves similar result to NARS at the first stage. With enhanced train set, the performance improves significantly. NARS\_SAGN+2-SLE has the mean test accuracy of 54.40\%, which is state-of-the-art result.

\subsection{Runtime experiments}
We also conduct extra runtime experiments of different stages on ogbn-products. As shown in Table \ref{tab: runtime experiments of different stages}, the training runtime increases with stage increasing. Because the enhanced train set contains much more nodes. However the model will converge faster. 

We further report runtime experiments of SIGN and SAGN on ogbn-papers100M in \ref{tab: runtime experiments on ogbn-papers100M}. They achieve close results. In total, SAGN has less GPU memory cost and parameters. While SAGN has a bit longer inference runtime.

\begin{table}[htb]
\caption{Runtimes experiments of different stages on ogbn-products. With additional nodes with pseudo labels, the training process costs more time. While the model converges in less epochs.}\label{tab: runtime experiments of different stages}
\begin{center}
    \begin{tabular}{cccc}
       \toprule
       Method & Training & Inference & Memory \\
       \midrule
       SAGN+0-SLE & 1.47±0.01s         & 8.35±0.33s & 14894Mb\\
       SAGN+1-SLE & 14.95±0.08s        & 8.42±0.39s & 14894Mb\\
       SAGN+2-SLE & 17.01±0.18s        & 8.34±0.39s & 14894Mb\\
       \bottomrule
    \end{tabular}
\end{center}
\end{table}

\begin{table}[tb]
\caption{Runtime experiments on ogbn-papers100M. The runtimes (seconds) with means and standard deviations, memory costs (Mb) and parameter numbers over 10 runs are reported.}\label{tab: runtime experiments on ogbn-papers100M}
\begin{center}
    \begin{tabular}{ccccc}
    \toprule
       Method & Training & Inference & Memory & Parameters\\
    \midrule
       SIGN         & 11.72±0.31s & 6.77±0.49s  & 10700Mb & 9106610\\
       SAGN         & 11.79±1.37s  & 6.84±0.49s & 7484Mb & 6098092\\
       \midrule
       SIGN+0-SLE  & 14.86±0.41s   & 6.67±0.74s & 10770Mb & 11565406\\
       SAGN+0-SLE & 14.31±1.22s   & 7.05±0.68s & 7842Mb & 8556888\\

    \bottomrule
    \end{tabular}
\end{center}
\end{table}

\begin{figure}[htb]
\centering

\includegraphics[width=\linewidth]{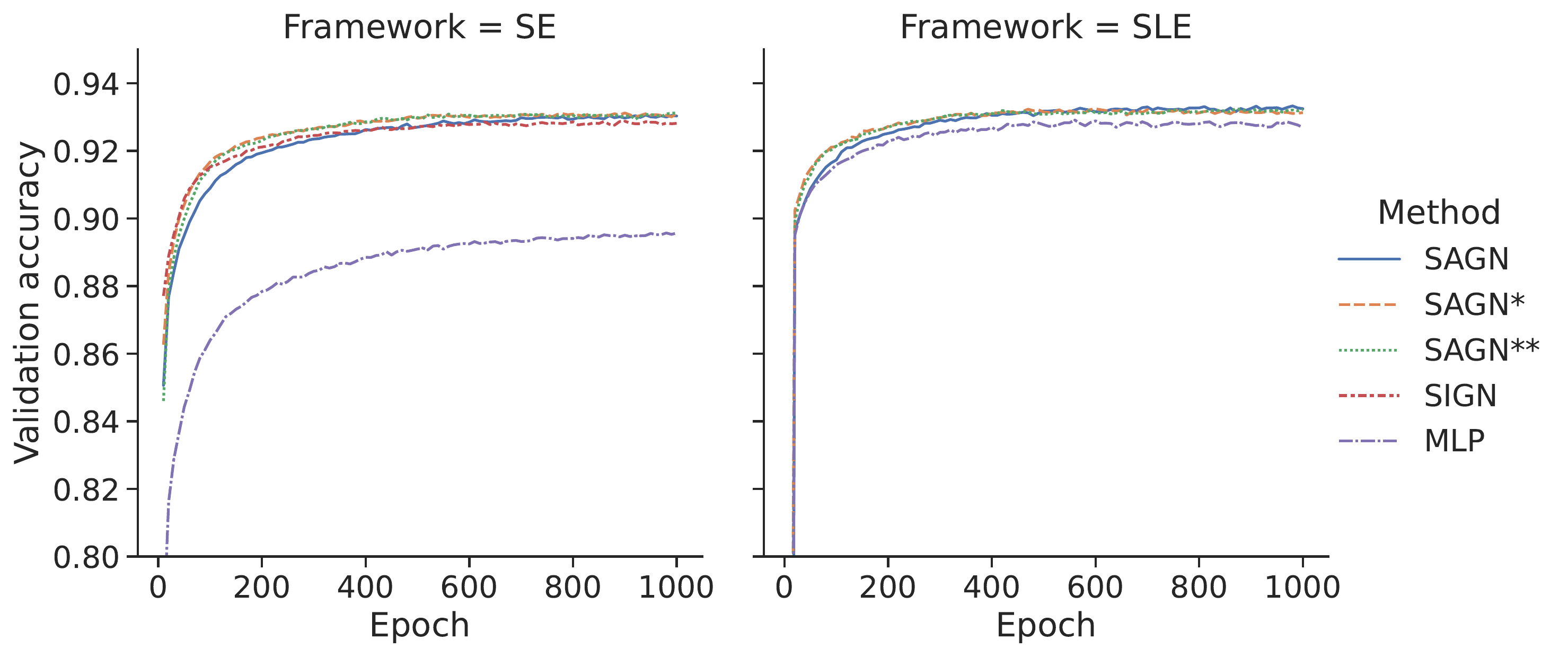}

\centering
\caption{Converge statistics of different base models under SE and SLE at the first stage. The validation accuracy for every 10 epochs is reported.}
\label{fig: base models converge}
\end{figure}

\begin{figure}[htb]
\centering
\includegraphics[width=\linewidth]{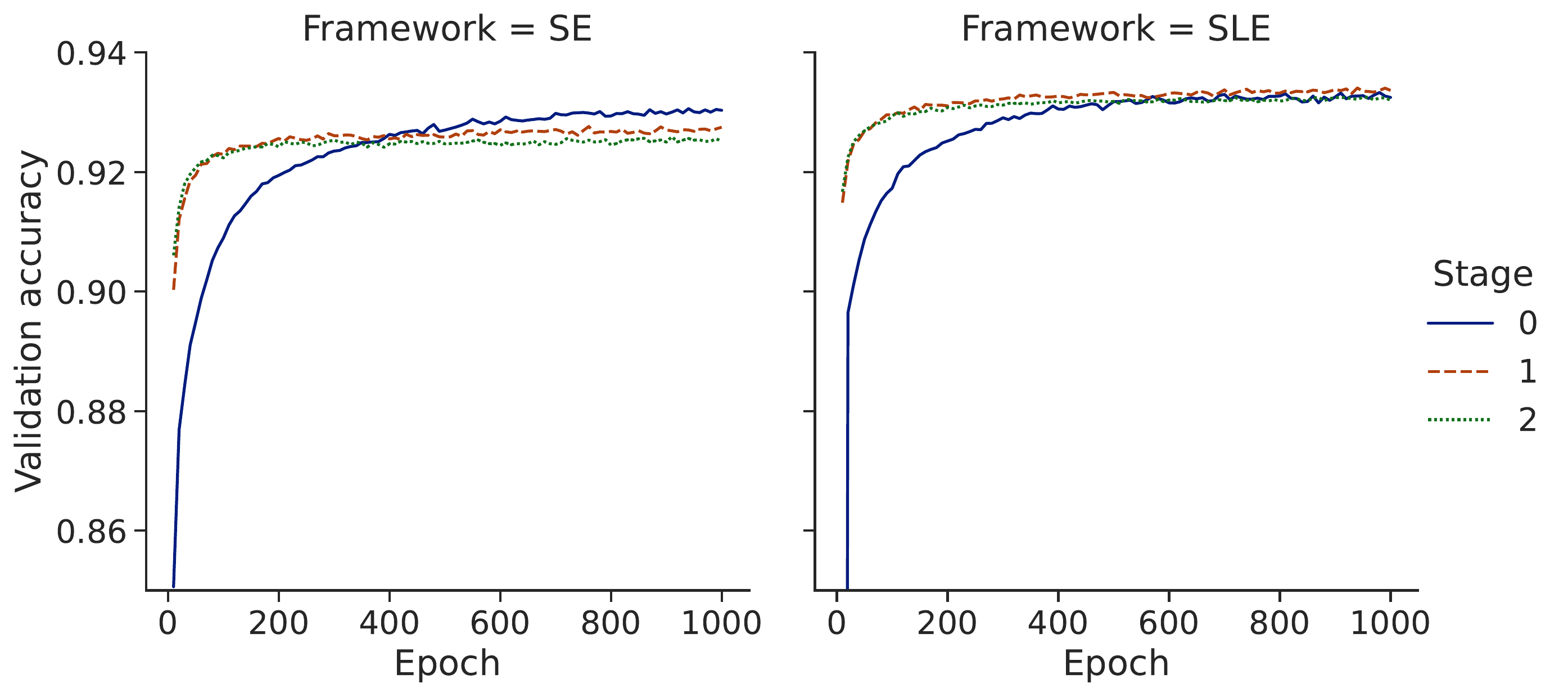}
\caption{Converge statistics of SAGN under SE and SLE at 3 stages.}
\label{fig: SAGN converge}
\end{figure}

\begin{figure}[htb]        
\centering
\includegraphics[width=\linewidth]{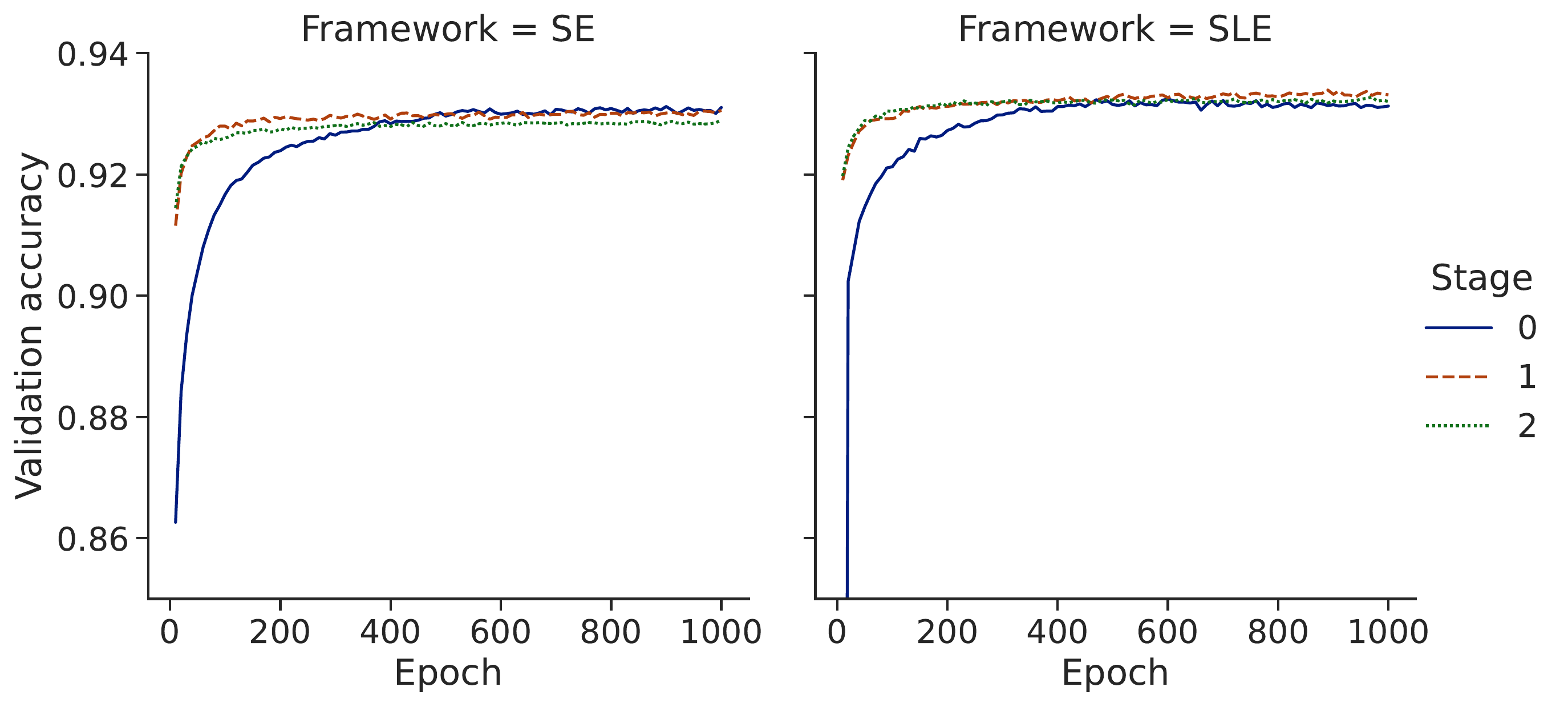}
\caption{Converge statistics of SAGN* under SE and SLE at 3 stages.}
\label{fig: SAGN* converge}
\end{figure}

\begin{figure}[htb]
\centering
\includegraphics[width=\linewidth]{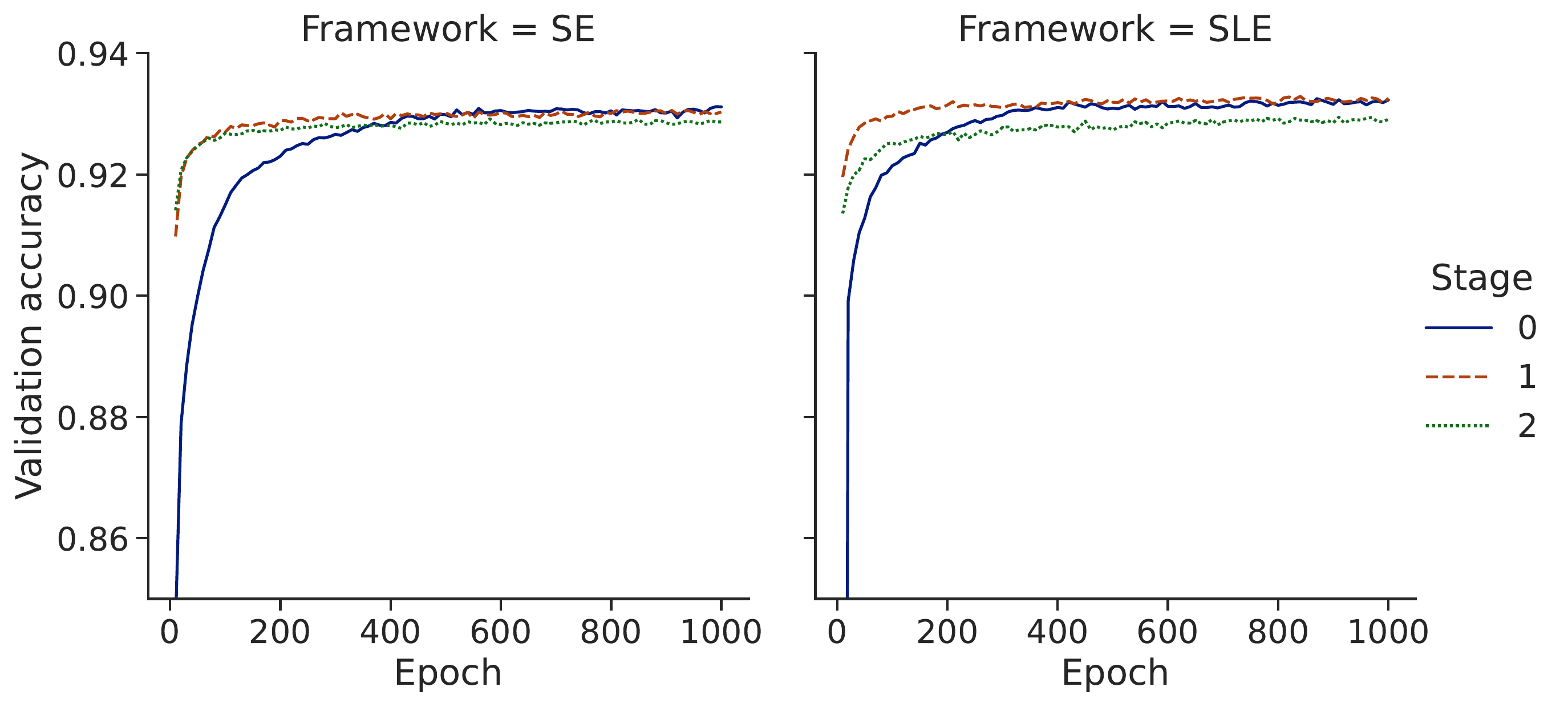}
\caption{Converge statistics of SAGN** under SE and SLE at 3 stages.}
\label{fig: SAGN** converge}
\end{figure}

\begin{figure}[htb]
\centering
\includegraphics[width=\linewidth]{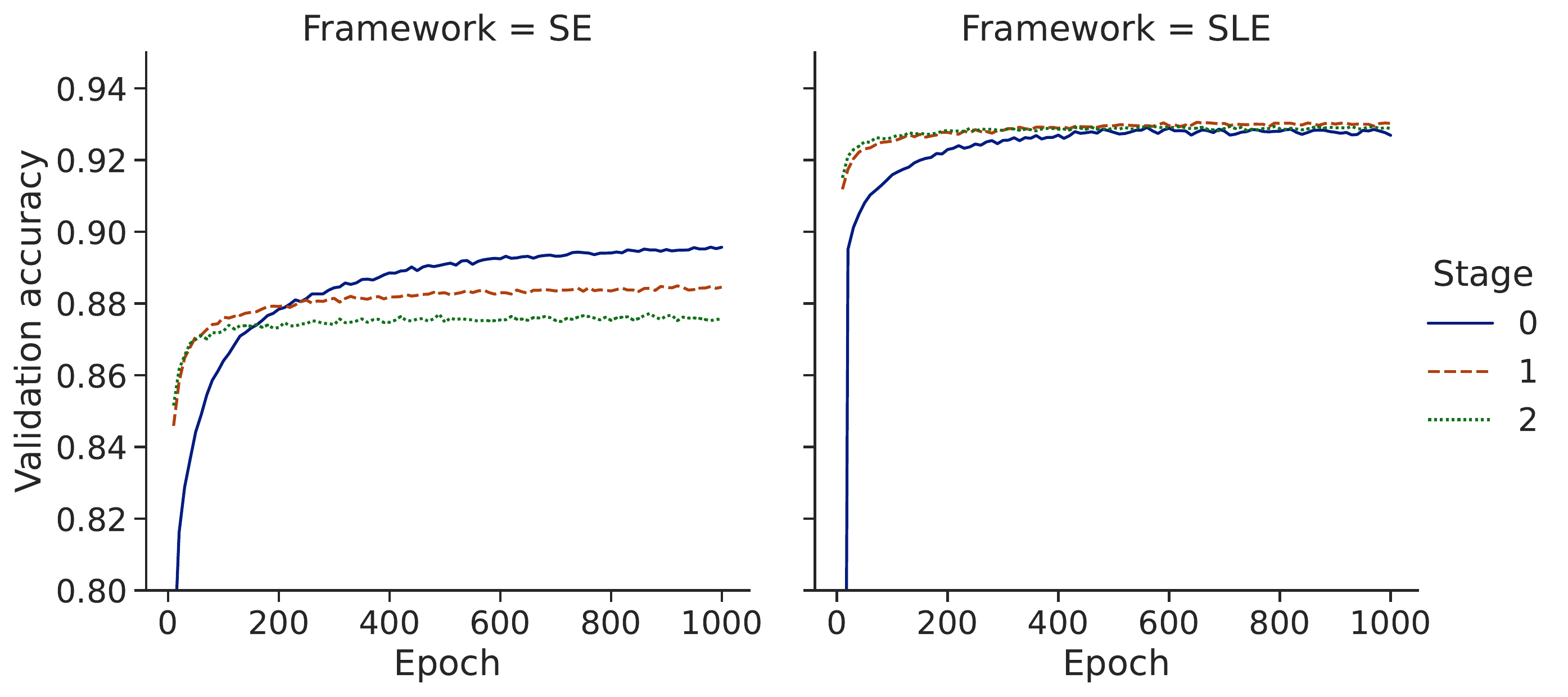}
\caption{Converge statistics of MLP under SLE and SE at 3 stages.}
\label{fig: MLP converge}
\end{figure}

\begin{figure}[htb]
\centering
\includegraphics[width=0.5\linewidth]{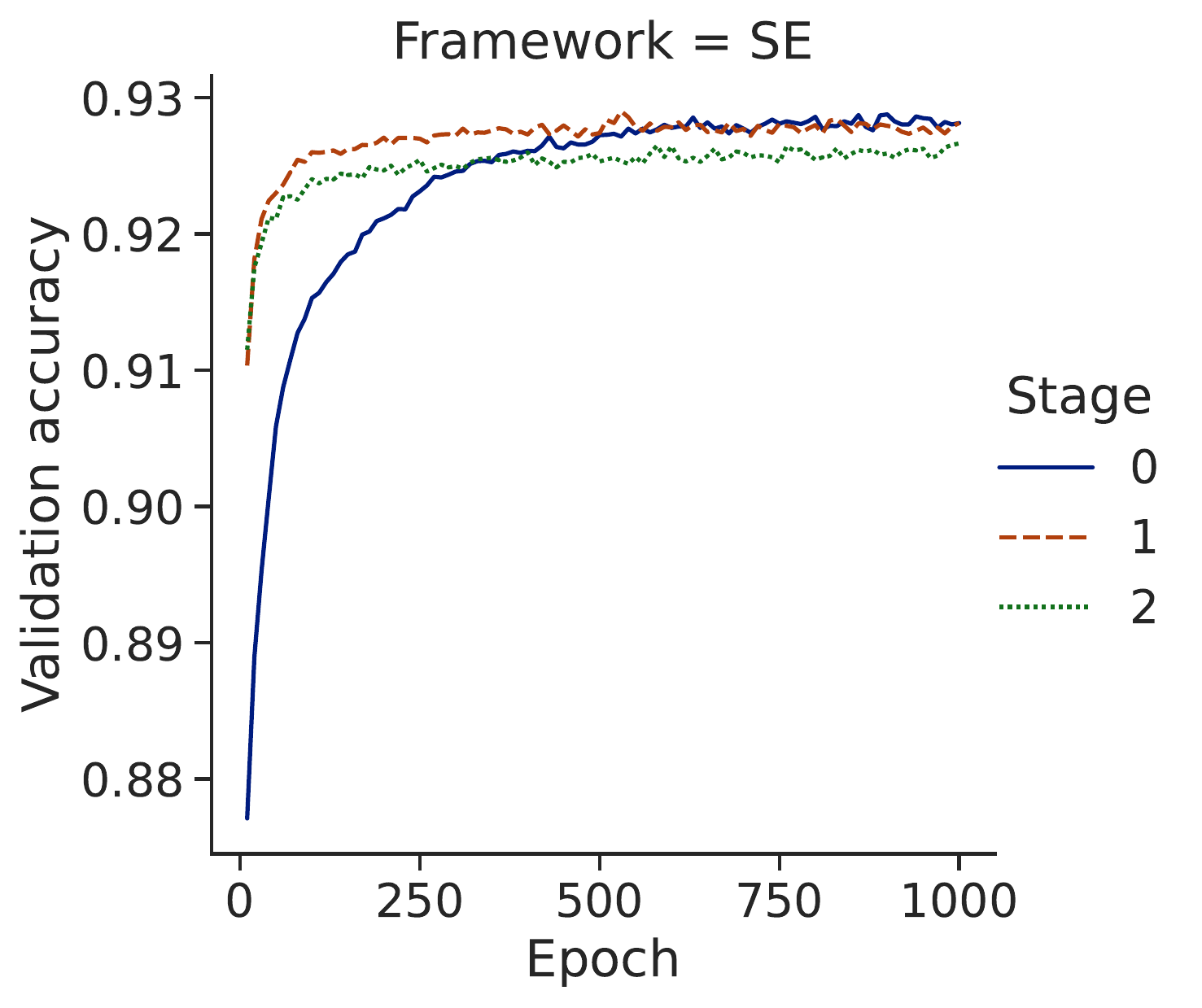}
\caption{Converge statistics of SIGN under SE at 3 stages. SIGN+SLE is not reported due to OOM error.}
\label{fig: SIGN converge}
\end{figure}

\section{Converge analysis}
We record validation accuracies of different base models under SE and SLE at three stages on ogbn-products for every 10 epochs. As shown in Figure \ref{fig: base models converge}, with label model, all models converge faster and have similar converge curves. As shown in Figures \ref{fig: SAGN converge}, \ref{fig: MLP converge}, \ref{fig: SAGN* converge}, \ref{fig: SAGN** converge} and \ref{fig: SIGN converge}, models converge faster with enhanced train set. Because the model at later stages can benefit from previous model's output. To control training time without losing performance, we cutoff training models in earlier epochs at later stages in actual usage. A special case is that MLP cannot be improved by self-training without label model. Although it still converges faster with enhanced train set. In constrat, MLP with label model can be improved by self-training.

\paragraph{Overfitting from validation set}
With stages increasing, validation and test accuracies of SAGN are improved on ogbn-papers100M and ogbn-mag while validation accuracies become a bit lower on ogbn-products. We think that very low validation ratio (2\%) in ogbn-products results in this phenomenon. Because we select best model based on results of very few nodes from validation test. With enhanced train set containing nodes from both validation and test sets, the overfitting from best model selection on validation set is alleviated. Thus validation and test accuracies are more close and overall results are improved. In details, the validation accuracy is reduced and test accuracy is increased.

\begin{table}[htb]
\caption{Hyperparameter settings of SAGN+SLE for inductive datasets. At the first stage we do not use label model.}\label{tab: hyperparameter settings inductive}
    \centering
    \setlength{\tabcolsep}{1mm}{
    \begin{tabular}{ccccccccccccc}
    \toprule
         Dataset & $lr$ & $hidden$ & $norm$ & $L$ & $K_f$ & $K_l$ & $drop$ & $drop_{a}$ & $drop_{i}$ & $batch\_size$ & $a$ & $wd$\\
    \midrule
         Reddit & 1e-4 & 512 & row & 2 & 2 & 4 & 0.7 & 0.0 & 0.0 & 1000 & 0.9 & 0\\
         Flickr & 1e-3 & 512 & sym & 2 & 2 & 2 & 0.7 & 0.0 & 0.0 & 256 & 0.5 & 3e-6 \\
         PPI & 1e-3 & 1024 & row & 2 & 2 & 9 & 0.3 & 0.1 & 0.0 & 256 & 0.9 & 3e-6 \\
         Yelp & 1e-4 & 256 & row & 2 & 2 & 6 & 0.05 & 0.0 & 0.0 & 200 & 0.9 & 5e-6 \\
    \bottomrule
    \end{tabular}
    }
\end{table}

\begin{table}[htb]
\caption{Hyperparameter settings for ogbn-products including ablation studies.}\label{tab: hyperparameter settings ogbn-products}
    \centering
    \setlength{\tabcolsep}{1mm}{
    \begin{tabular}{ccccccccccccc}
    \toprule
         Method & $lr$ & $hidden$ & $norm$ & $L$ & $K_f$ & $K_l$ & $drop$ & $drop_{a}$ & $drop_{i}$ & $batch\_size$ & $a$ & $wd$ \\
    \midrule
         MLP+SE & 1e-3 & 512 & row & 4 & 9 & - & 0.5 & - & 0.2 & 50000 & 0.9 & 0 \\
         SIGN+SE & 1e-3 & 512 & row & 2 & 5 & - & 0.4 & - & 0.3 & 50000 & 0.9 & 0 \\
         SAGN*+SE & 1e-3 & 512 & row & 2 & 5 & - & 0.5 & - & 0.2 & 50000 & 0.9 & 0 \\
         SAGN**+SE & 1e-3 & 512 & row & 2 & 5 & - & 0.5 & - & 0.2 & 50000 & 0.9 & 0 \\
         SAGN+SE & 1e-3 & 512 & row & 2 & 5 & - & 0.5 & 0.4 & 0.2 & 50000 & 0.9 & 0\\
         \midrule
         SLE & 1e-3 & 512 & row & 4 & - & 9 & 0.5 & - & - & 50000 & 0.9 & 0 \\
         MLP+SLE & 1e-3 & 512 & row & 4 & 9 & 9 & 0.5 & - & 0.2 & 50000 & 0.9 & 0 \\
         SIGN+SLE & 1e-3 & 512 & row & 2 & 5 & 9 & 0.4 & - & 0.3 & 50000 & 0.9 & 0 \\
         SAGN*+SLE & 1e-3 & 512 & row & 2 & 5 & 9 & 0.5 & - & 0.2 & 50000 & 0.9 & 0 \\
         SAGN**+SLE & 1e-3 & 512 & row & 2 & 5 & 9 & 0.5 & - & 0.2 & 50000 & 0.9 & 0 \\
         SAGN+SLE & 1e-3 & 512 & row & 2 & 3 & 9 & 0.5 & 0.4 & 0.2 & 50000 & 0.9 & 0 \\
    \bottomrule
    \end{tabular}
    }
\end{table}

\begin{table}[htb]
\caption{Hyperparameter settings for ogbn-papers100M.}\label{tab: hyperparameter settings ogbn-papers100M}
    \centering
    \setlength{\tabcolsep}{1mm}{
    \begin{tabular}{ccccccccccccc}
    \toprule
         Method & $lr$ & $hidden$ & $norm$ & $L$ & $K_f$ & $K_l$ & $drop$ & $drop_{a}$ & $drop_{i}$ & $batch\_size$ & $a$ & $wd$\\
    \midrule
         SAGN+SE & 1e-3 & 1024 & row & 2 & 3 & - & 0.4 & 0.0 & 0.0 & 5000 & 0.7 & 0\\
         SAGN+SLE & 1e-3 & 1024 & row & 2 & 3 & 9 & 0.4 & 0.0 & 0.0 & 5000 & 0.7 & 0 \\
    \bottomrule
    \end{tabular}
    }
\end{table}

\begin{table}[htb]
\caption{Hyperparameter settings for ogbn-mag.}\label{tab: hyperparameter settings ogbn-mag}
    \centering
    \setlength{\tabcolsep}{1mm}{
    \begin{tabular}{ccccccccccccc}
    \toprule
         Method & $lr$ & $hidden$ & $norm$ & $L$ & $K_f$ & $K_l$ & $drop$ & $drop_{a}$ & $drop_{i}$ & $batch\_size$ & $a$ & $wd$\\
    \midrule
         SAGN+SLE & 2e-3 & 512 & row & 2 & 5 & 3 & 0.3 & 0.2 & 0.0 & 50000 & 0.4 & 0 \\
         SAGN+TransE+SLE & 2e-3 & 512 & row & 2 & 5 & 3 & 0.5 & 0.0 & 0.0 & 50000 & 0.4 & 0 \\
         NARS\_SAGN+SLE & 1e-3 & 512 & row & 2 & 5 & 3 & 0.5 & 0.0 & 0.0 & 10000 & 0.4 & 0 \\
    \bottomrule
    \end{tabular}
    }
\end{table}

\section{Hyperparameter settings}
For hyperparameters, we simply tune them empirically. Let $lr$ represent learning rate, $hidden$ represent hidden dimension, $norm$ represent whether use symmetrically normalized adjacency matrix (sym) or row-stochastic random walk matrix (row), $L$ represent layer number in MLP encoders, $K_f$ represent hop number of node feature aggregations, $K_l$ represent hop number of node label aggregations, $drop$ represent dropout ratio, $drop_a$ represent attention weights dropout ratio, $drop_i$ represent input dropout ratio, $batch\_size$ represent batch size in training process, $a$ represent threshold for filtering confident nodes and $wd$ represent weight decay ratio. The hyperparameter settings for results in different datasets are shown in Table \ref{tab: hyperparameter settings inductive}, Table \ref{tab: hyperparameter settings ogbn-products}, Table \ref{tab: hyperparameter settings ogbn-papers100M} and Table \ref{tab: hyperparameter settings ogbn-mag}. Note that the settings of epoch numbers are omitted in tables. For Reddit, the epoch numbers are set to $[500, 500, 500]$. For Flickr, the epoch numbers are set to $[50, 50, 50]$. For PPI, the epoch numbers are set to $[2000, 2000, 2000]$. For Yelp, the epoch numbers are set to $[100, 100, 100]$. For ogbn-products, all methods use epoch numbers of $[1000, 200, 200]$ except that MLP uses epoch numbers of $[1000, 500, 500]$. For ogbn-papers100M, the epoch numbers are set to $[100, 50, 50]$. For SAGN on ogbn-mag, the epoch numbers are set to $[200, 200, 200]$. For NARS\_SAGN on ogbn-mag, the epoch numbers are set to $[500, 200, 150]$.

\paragraph{Ablation study}
SIGN, SAGN and its variants share the same MLP layer number of 2, hop number of 5 and hidden dimension of 512 except SAGN+SLE which uses a smaller hop number of 3.

\paragraph{Runtime experiments}
For runtime experiments on ogbn-products, the most of hyperparameters are kept the same. The batch sizes for training and inference are respectively set to 50000 and 100000 for all methods. The hidden dimensions of all methods are set to 512. The number of layers in MLP is 4. In both SIGN and SAGN, the number of layers in multi-hop MLP encoders is 2 and the number of post MLP encoder is 2. The number of aggregations in SIGN and SAGN is set to 5. 0-SLE means incorporating label model. SIGN+0-SLE will encounter OOM error.

Similarly, for runtime experiments on ogbn-papers100M, all hyperparameters associated to runtime and memory cost are the same. The batch sizes for training and inference are respectively set to 5000 and 100000. The hidden dimensions are set to 512. The number of layers in every MLP encoder is set to 2. The number of aggregations is set to 3.

\section{Visualization}
\begin{figure}[tb]
    \centering
    \subfigure[SAGN+0-SE]{
    \includegraphics[width=0.45\linewidth, trim=0 0 50 0, clip]{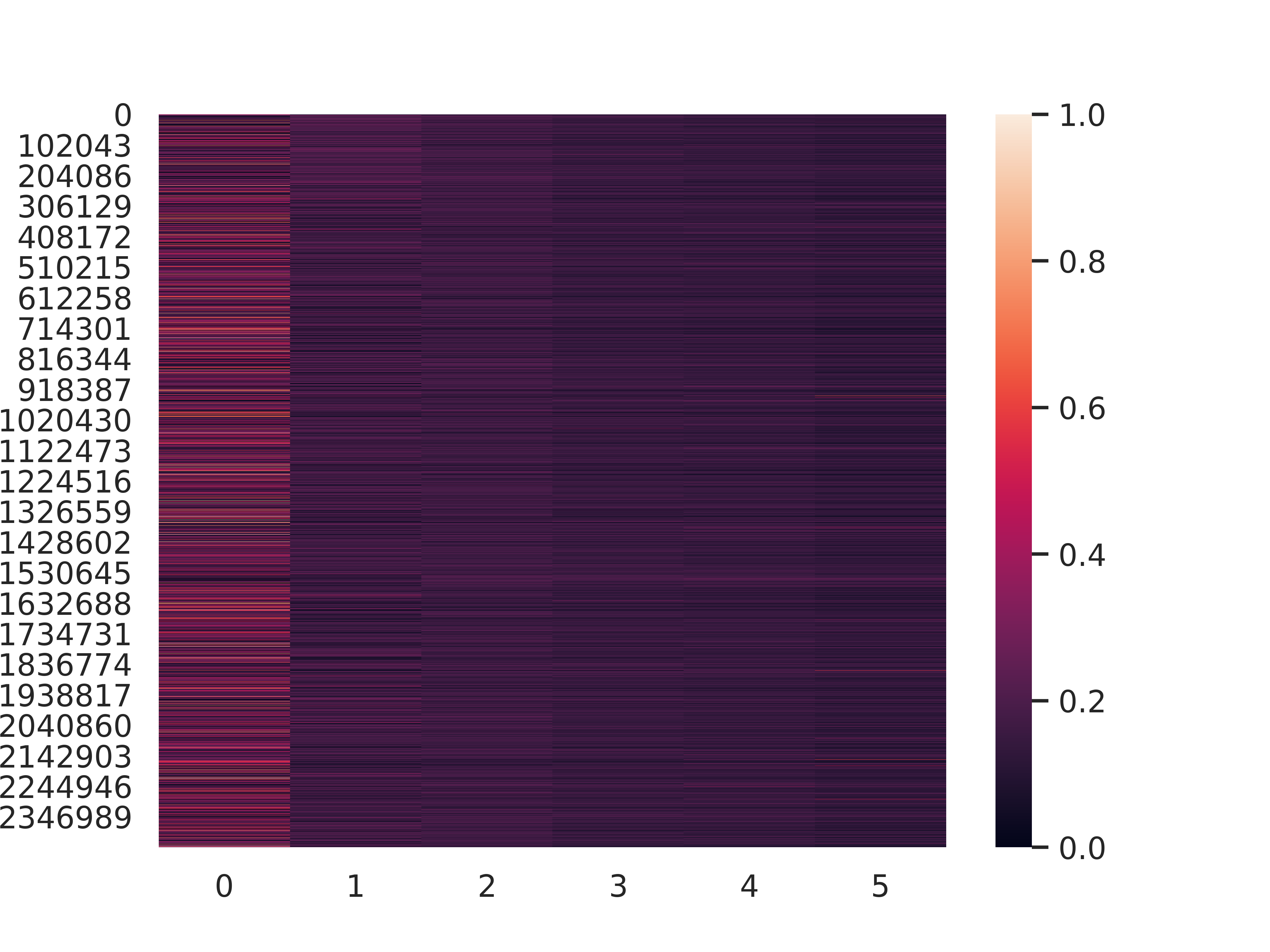}
    \label{fig: attn_ogbn_products_SE_0}
    }
    \subfigure[SAGN+0-SLE]{
    \includegraphics[width=0.45\linewidth, trim=0 0 50 0, clip]{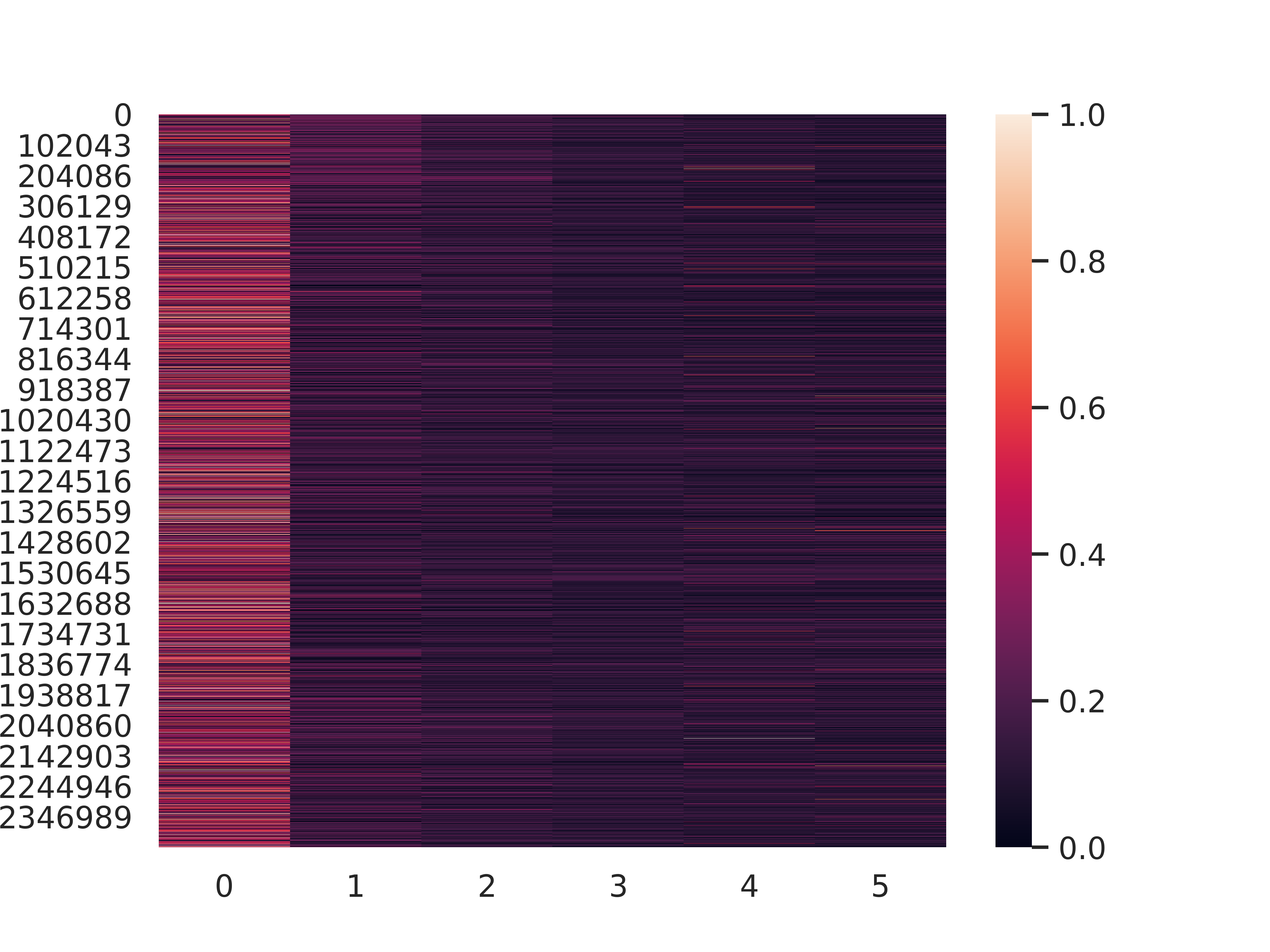}
    \label{fig: attn_ogbn_products_SLE_0}
    }
    
    \subfigure[SAGN+1-SE]{
    \includegraphics[width=0.45\linewidth, trim=0 0 50 0, clip]{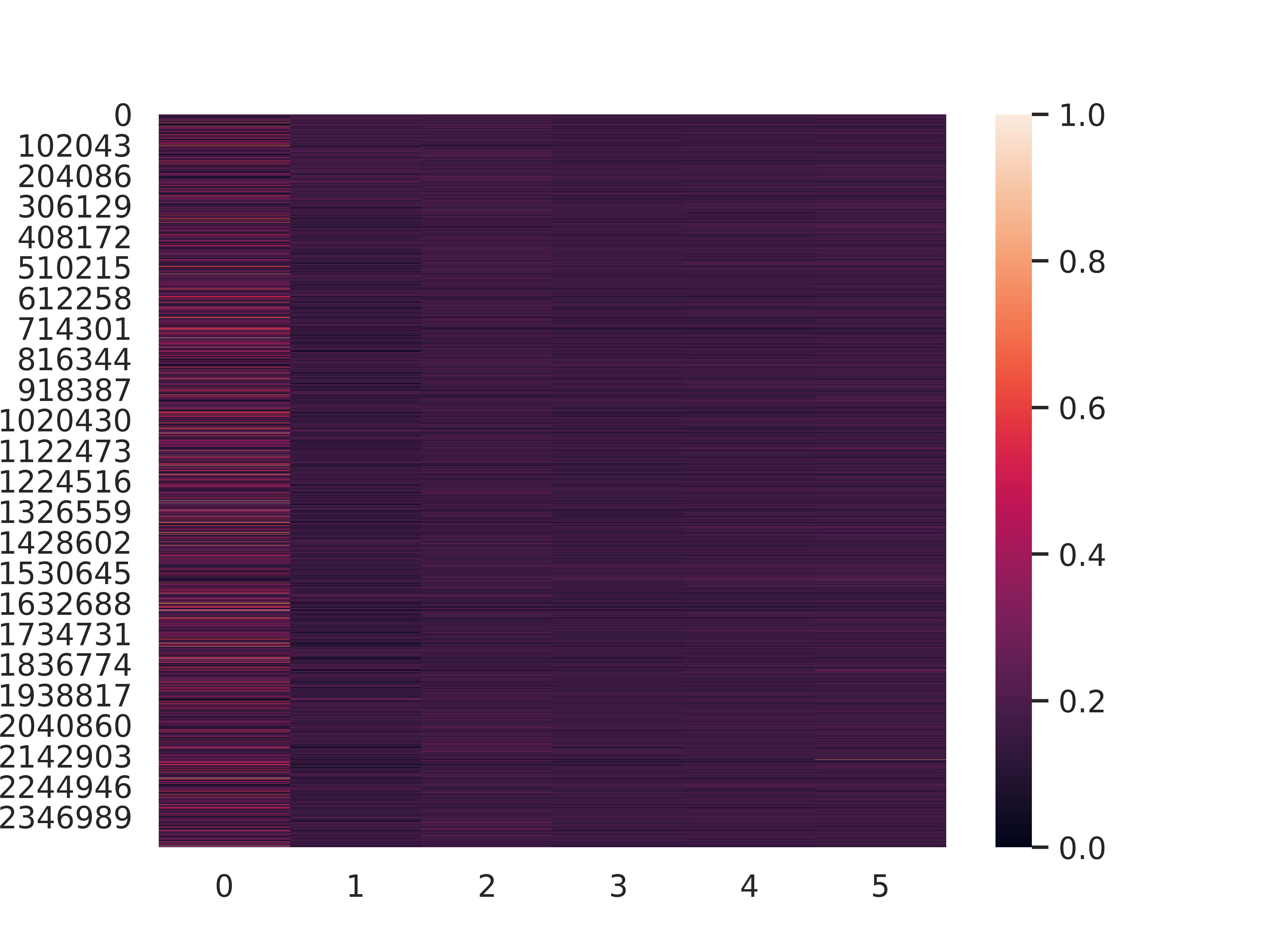}
    \label{fig: attn_ogbn_products_SE_1}
    }
    \subfigure[SAGN+1-SLE]{
    \includegraphics[width=0.45\linewidth, trim=0 0 50 0, clip]{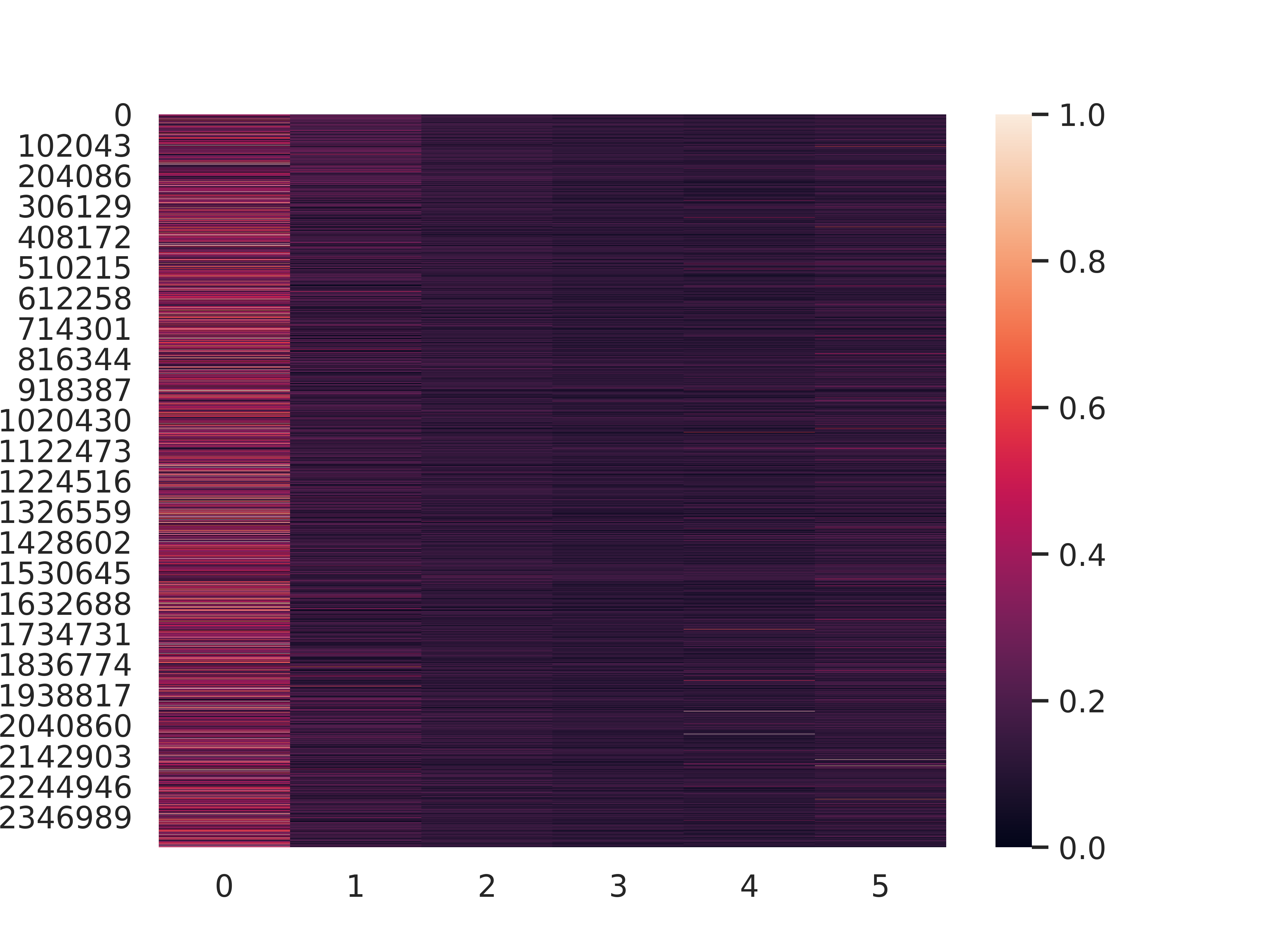}
    \label{fig: attn_ogbn_products_SLE_1}
    }
    
    \subfigure[SAGN+2-SE]{
    \includegraphics[width=0.45\linewidth, trim=0 0 50 0, clip]{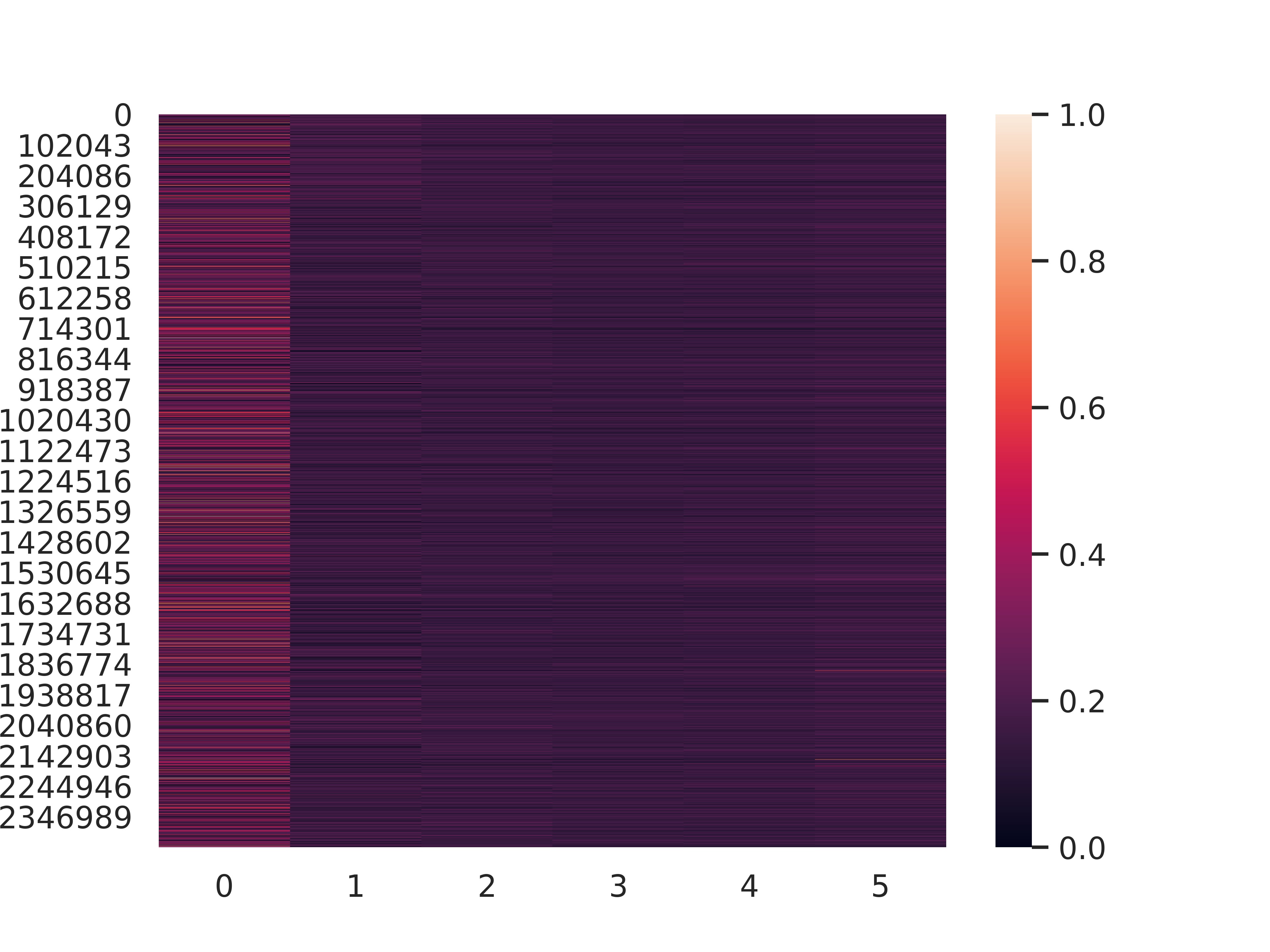}
    \label{fig: attn_ogbn_products_SE_2}
    }
    \subfigure[SAGN+2-SLE]{
    \includegraphics[width=0.45\linewidth, trim=0 0 50 0, clip]{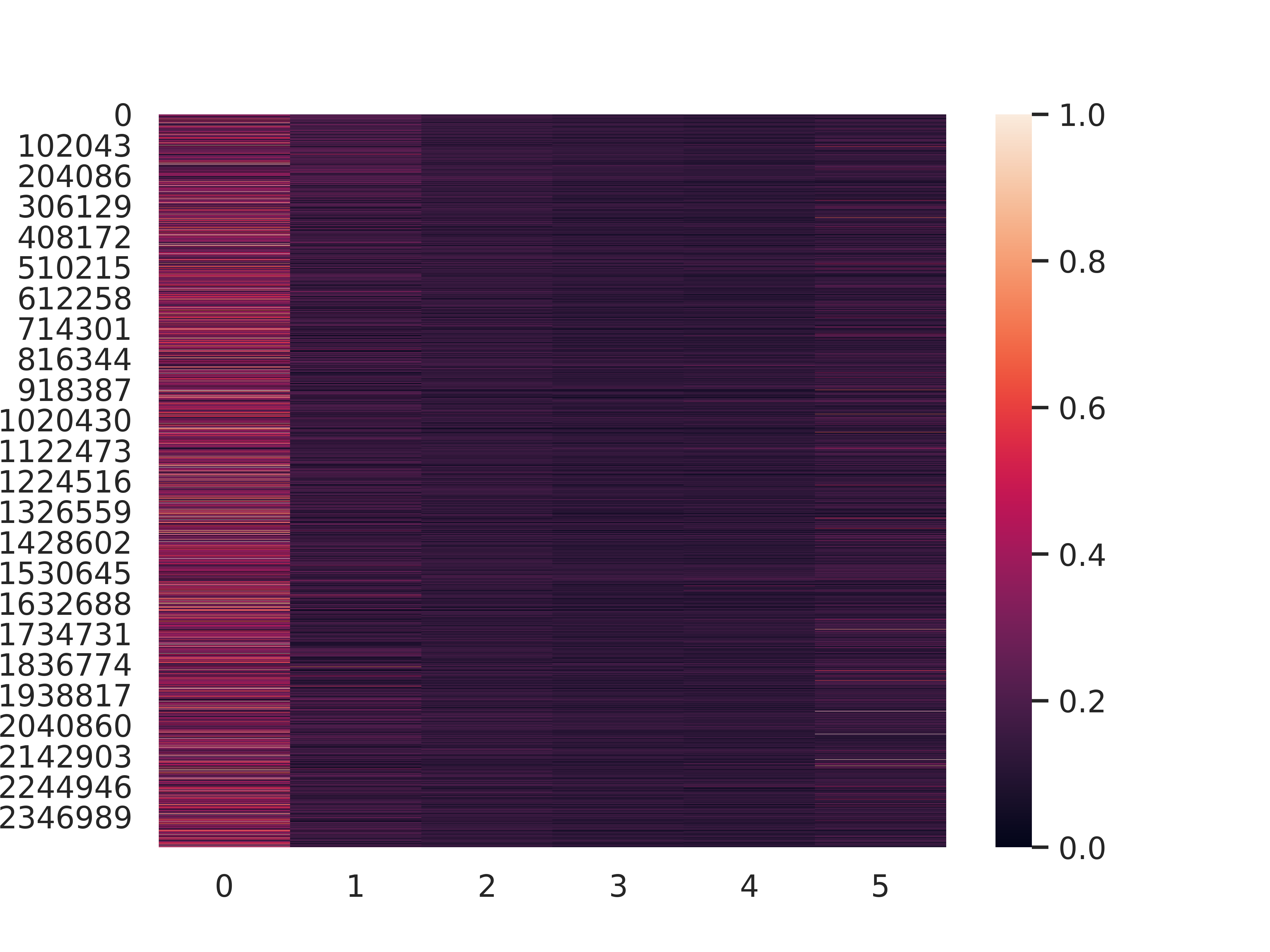}
    \label{fig: attn_ogbn_products_SLE_2}
    }

    \caption{Heat map of attention weights. SAGN+SE and SAGN+SLE at different stages on ogbn-products are reported. The horizontal axis represents hop number and the vertical axis represents node id.}
    \label{fig: attn_ogbn_products}
\end{figure}

\begin{figure}[tb]
    \centering
    \subfigure[SAGN+0-SE]{
    \includegraphics[width=0.45\linewidth, trim=0 0 50 0, clip]{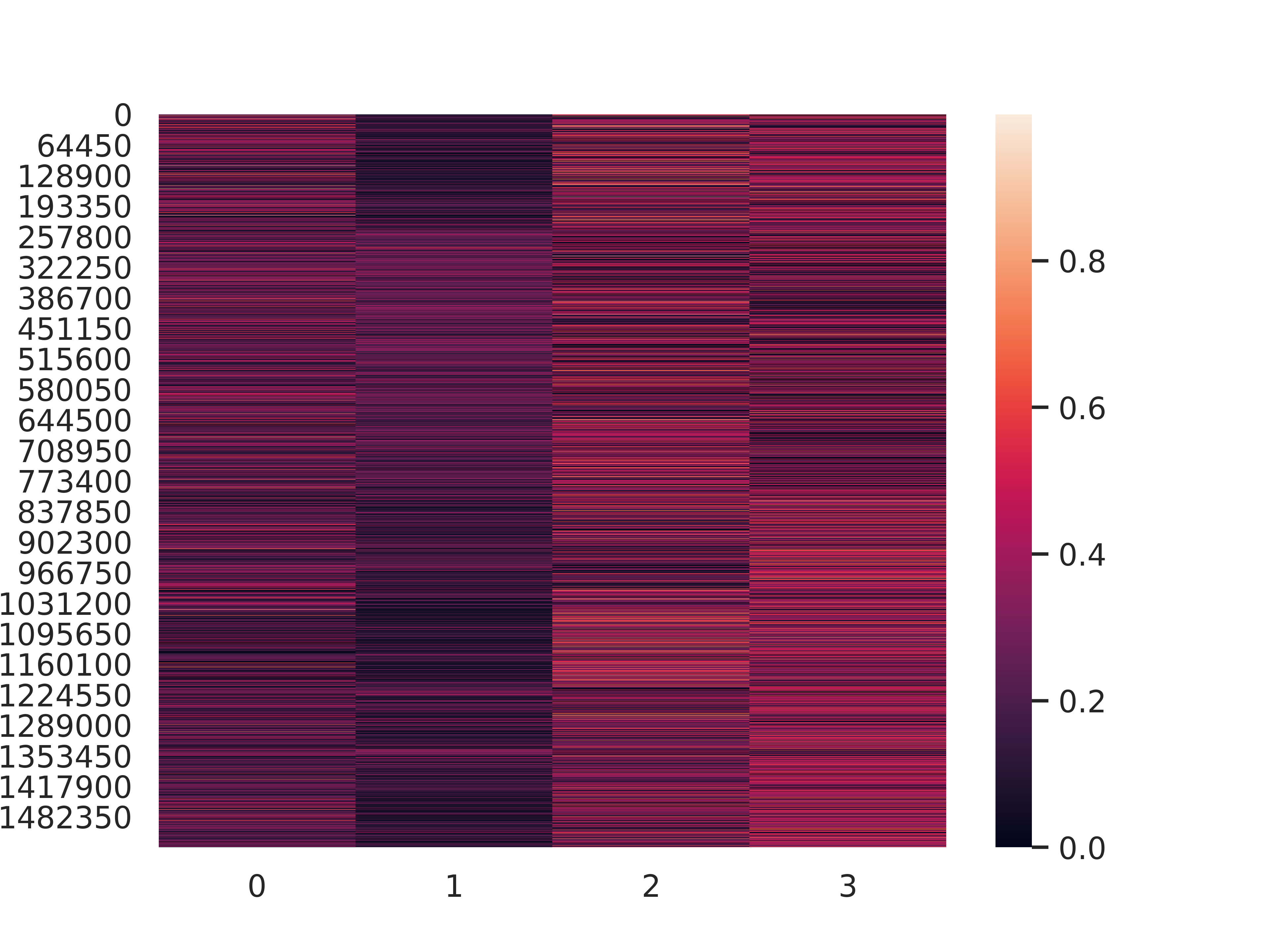}
    \label{fig: attn_ogbn_papers100M_SE_0}
    }
    \subfigure[SAGN+0-SLE]{
    \includegraphics[width=0.45\linewidth, trim=0 0 50 0, clip]{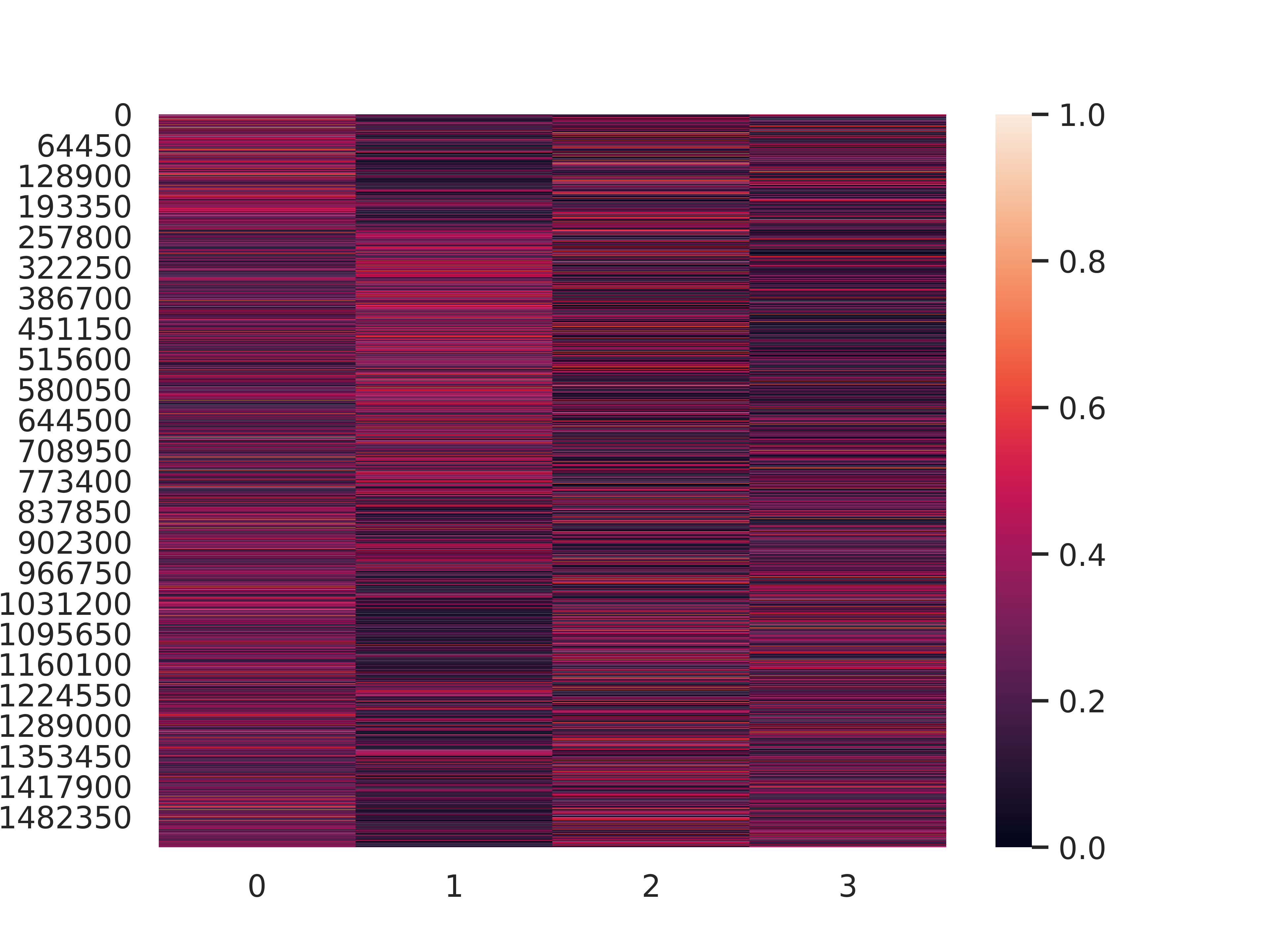}
    \label{fig: attn_ogbn_papers100M_SLE_0}
    }
    
    \subfigure[SAGN+1-SE]{
    \includegraphics[width=0.45\linewidth, trim=0 0 50 0, clip]{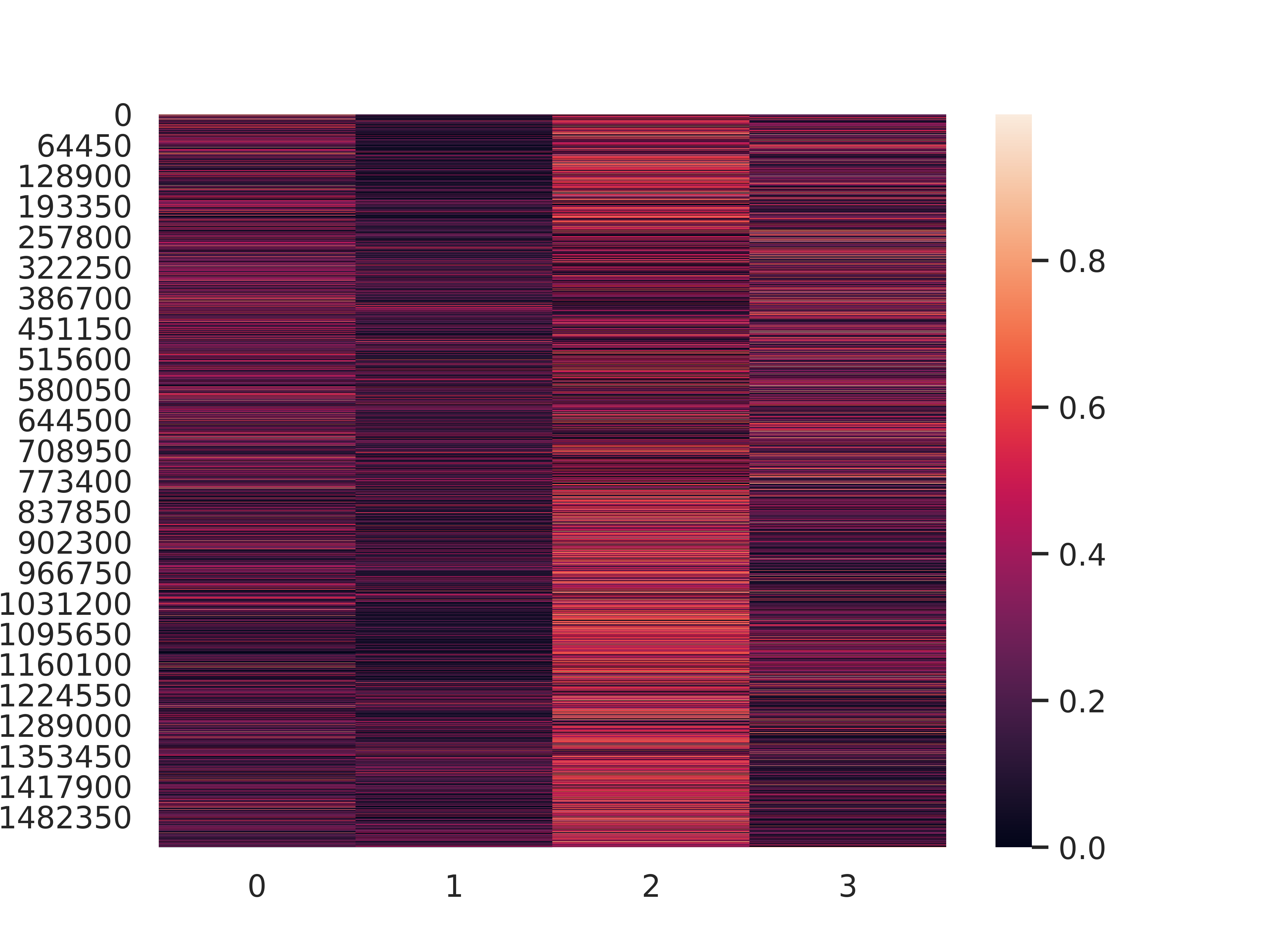}
    \label{fig: attn_ogbn_papers100M_SE_1}
    }
    \subfigure[SAGN+1-SLE]{
    \includegraphics[width=0.45\linewidth, trim=0 0 50 0, clip]{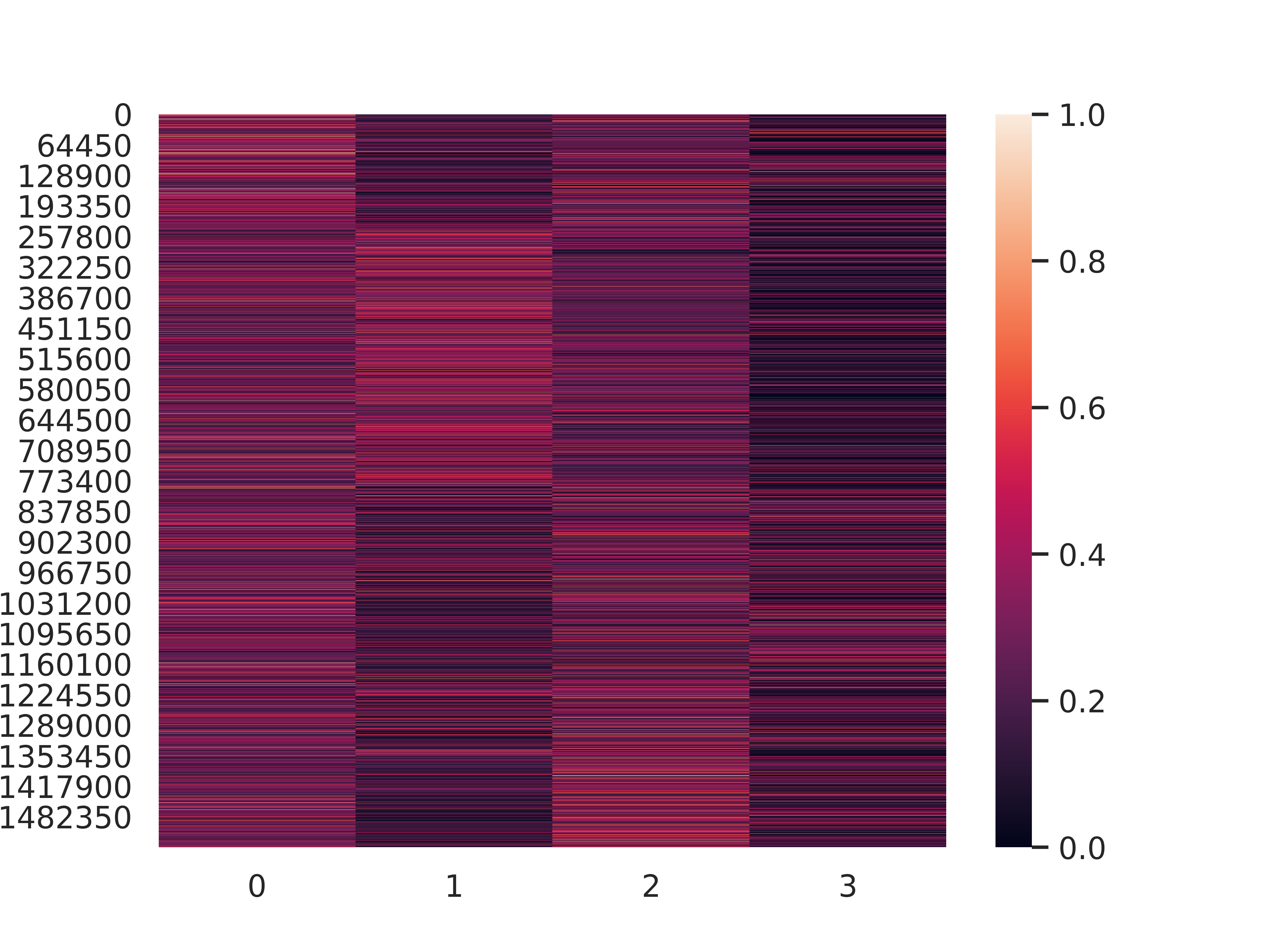}
    \label{fig: attn_ogbn_papers100M_SLE_1}
    }
    
    \subfigure[SAGN+2-SE]{
    \includegraphics[width=0.45\linewidth, trim=0 0 50 0, clip]{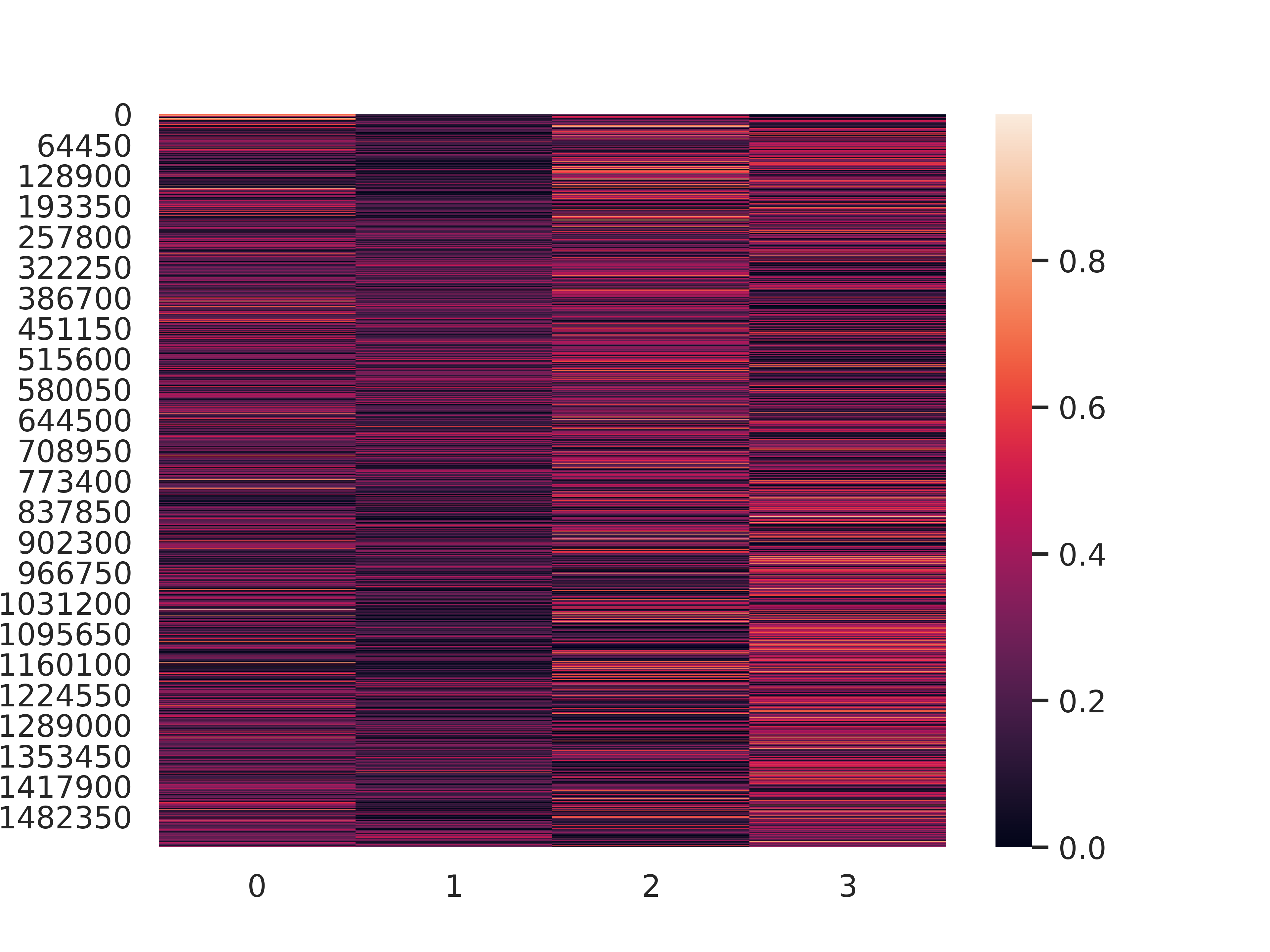}
    \label{fig: attn_ogbn_papers100M_SE_2}
    }
    \subfigure[SAGN+2-SLE]{
    \includegraphics[width=0.45\linewidth, trim=0 0 50 0, clip]{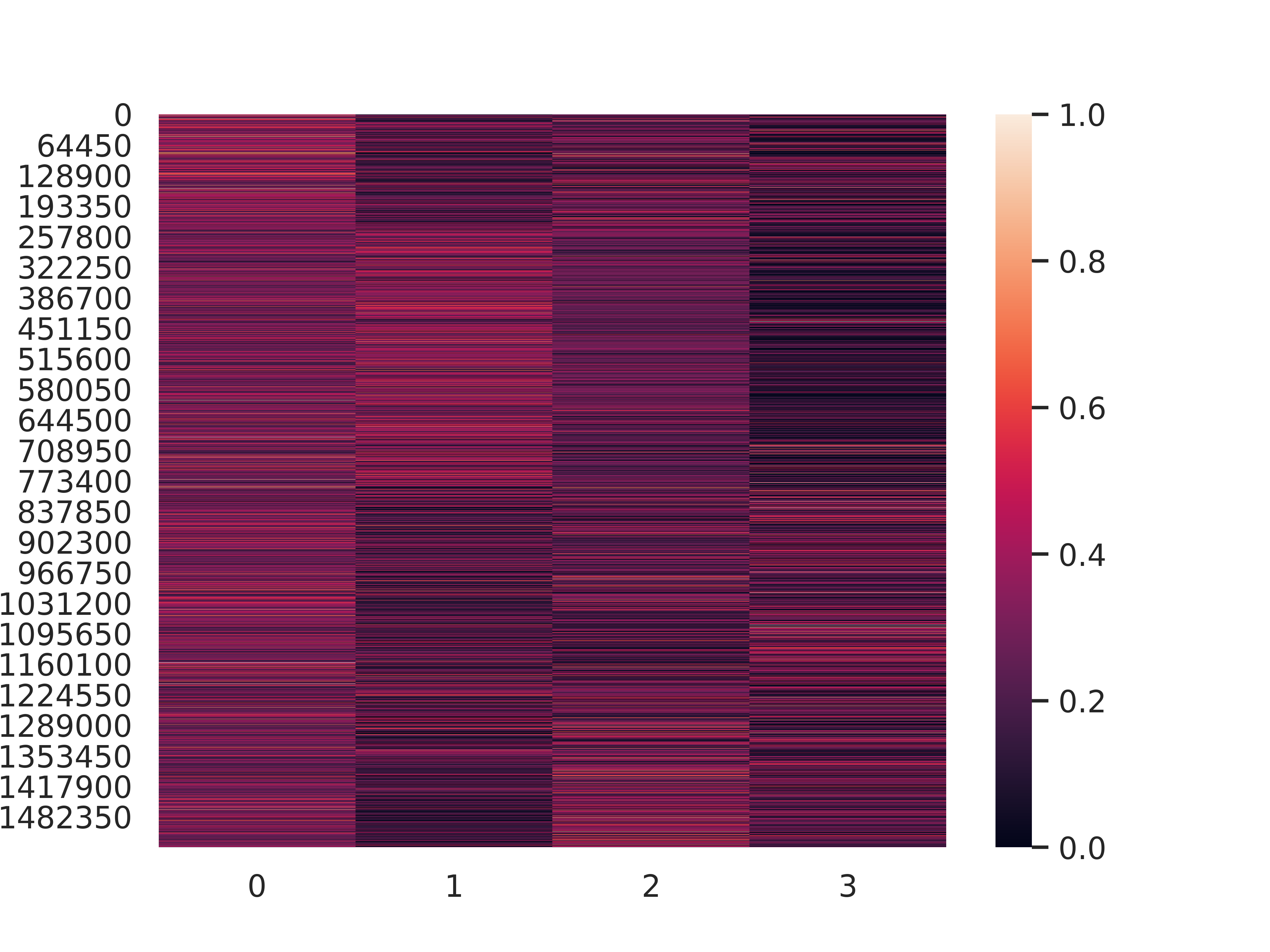}
    \label{fig: attn_ogbn_papers100M_SLE_2}
    }

    \caption{Heat map of attention weights. SAGN+SE and SAGN+SLE at different stages on ogbn-papers100M are reported. The horizontal axis represents hop number and the vertical axis represents node id.}
    \label{fig: attn_ogbn_papers100M}
\end{figure}

The heatmap visualization of attention weights shows the global and local importances of different hops. We take ogbn-products and ogbn-papers100M for examples. As shown in Figure \ref{fig: attn_ogbn_products}, we can view the distribution of attention weights on ogbn-products. Neither adding label model nor self-training affects the learned importances of hops. Compared to other hops, the initial (0-hop) feature has obviously larger average weight. But it can be also small for some nodes. This implies that the raw node feature already contains important information and smoothed features can offer supplementary information. But the condition is very different on ogbn-papers100M, as shown in Figure \ref{fig: attn_ogbn_papers100M}. We can find that 2-hop and 3-hop are more important. With label model, the importances become more homogeneous. SAGN+1-SE slightly changes the distribution compared with SAGN+0-SE. But it is restored for SAGN+2-SE. In addition, there exist obvious differences among nodes. This manifests as the different color blocks in the same column.

\end{appendix}

\end{document}